\pdfoutput=1
\documentclass[11pt]{article}

\usepackage[final]{acl}

\usepackage{times}
\usepackage{latexsym}

\usepackage[T1]{fontenc}
\usepackage[utf8]{inputenc}

\usepackage{microtype}

\usepackage{inconsolata}

\usepackage{graphicx}
\usepackage{booktabs}
\usepackage{multirow}
\usepackage{amsmath}
\usepackage{enumitem}
\usepackage{tikz}
\usepackage{pythonhighlight}
\usepackage{xcolor}
\usepackage{listings}

\definecolor{deepgreen}{RGB}{34,139,34}

\begin{document}
\title{Interactive-KBQA: Multi-Turn Interactions for Knowledge Base Question Answering with Large Language Models}

\author{
  Guanming Xiong \\
  Peking University \\
  Beijing, China \\
  \texttt{gm\_xiong@pku.edu.cn} \\\And
  Junwei Bao\thanks{Corresponding author.} \\
  Zuoyebang Education \\
  Technology Co., Ltd.\\
  Beijing, China \\
  \texttt{baojunwei001@gmail.com} \\\And
  Wen Zhao \\
  Peking University \\
  Beijing, China \\
  \texttt{zhaowen@pku.edu.cn}
  }

\maketitle
\begin{abstract}
This study explores the realm of knowledge base question answering (KBQA). 
KBQA is considered a challenging task, particularly in parsing intricate questions into executable logical forms. 
Traditional semantic parsing (SP)-based methods require extensive data annotations, which result in significant costs. 
Recently, the advent of few-shot in-context learning, powered by large language models (LLMs), has showcased promising capabilities. However, fully leveraging LLMs to parse questions into logical forms in low-resource scenarios poses a substantial challenge. 
To tackle these hurdles, we introduce Interactive-KBQA, a framework designed to generate logical forms through direct interaction with knowledge bases (KBs). 
Within this framework, we have developed three generic APIs for KB interaction. 
For each category of complex question, we devised exemplars to guide LLMs through the reasoning processes. 
Our method achieves competitive results on the WebQuestionsSP, ComplexWebQuestions, KQA Pro, and MetaQA datasets with a minimal number of examples (shots). 
Importantly, our approach supports manual intervention, allowing for the iterative refinement of LLM outputs. 
By annotating a dataset with step-wise reasoning processes, we showcase our model's adaptability and highlight its potential for contributing significant enhancements to the field.\footnote{Code and data are available at: \url{https://github.com/JimXiongGM/Interactive-KBQA}}
\end{abstract}

\section{Introduction}

\begin{figure}
  \centering
  \includegraphics[width=0.45\textwidth,page=1]{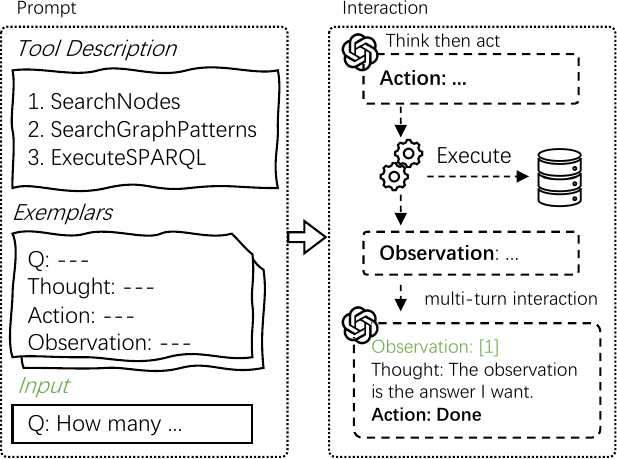}
  \caption{Overview of the interactive process.}
  \label{fig:model_overview}
\end{figure}

Knowledge base question answering (KBQA) is an increasingly significant research area that leverages structured knowledge bases (KBs) to provide precise answers to natural language (NL) questions. A KBQA system aims to harness the vast, structured information residing in KBs, such as Freebase \citep{Bollacker-Kurt-SIGMOD-2008-Freebase}, Wikidata \citep{Vrandečić-Denny-CACM-2014-Wikidata}, or domain-specific databases like the Movie KB \citep{Zhang-Yuyu-AAAI-2018-MetaQA}. Accurately interpreting and responding to user inquiries with data from knowledge bases has potent applications across numerous sectors, making it a focal point of both academic research and industrial innovation.

Recent advancements in KBQA research can generally be classified into two primary approaches: information retrieval (IR)-based methods and semantic parsing (SP)-based methods. IR-based methods focus on understanding the query, retrieving pertinent subgraphs from KBs related to the question, and extracting the answer from these subgraphs \citep{Dong-Guanting-CIKM-2023-Bridging-the-KB-Text-Gap, Zhang-Jing-ACL-2022-Subgraph-Retrieval-Enhanced-Model, Yan-Yuanmeng-EMNLP-2021-Large-scale-relation-learning}. Conversely, SP-based approaches convert NL questions into executable logical forms, leveraging pre-trained generative models to interact with the knowledge base and generate answers \citep{Chen-Shuang-ACL-2021-ReTraCk, Ye-Xi-ACL-2022-RNG-KBQA, Shu-Yiheng-EMNLP-2022-TIARA, Yu-Donghan-ICLR-2023-DecAF}.

The emergence of large language models (LLMs), such as ChatGPT \citep{Ouyang-Long-NeurIPS-2022-InstructGPT} and GPT-4 \citep{OpenAI-2023-GPT4}, has opened new avenues for enhancing KBQA systems. These models have demonstrated promising results in reasoning \citep{Wei-Jason-NeurIPS-2022-ChainOfThought} and few-shot learning capabilities \citep{Chen-Wenhu-EACL-2023-LLMTableReasoners}, setting new benchmarks in the KBQA domain \citep{Gu-Yu-ACL-2023-Pangu, Li-Tianle-ACL-2023-KB-BINDER, Jiang-EMNLP-2023-StructGPT, Jiashuo-Sun-ICLR-2024-Think-on-Graph, Li-Zhenyu-arXiv-2023-FlexKBQA}.

Despite these advancements, KBQA systems face several pressing challenges.

% IR Drawbacks
\textbf{Complex query handling}. The primary challenge for IR-based approaches is the difficulty in processing complex queries. For instance, questions characterized solely by entity types or concepts, alongside numerical constraints (e.g., "How many basketball players are taller than 2 meters?"), pose significant challenges. Such queries demand an understanding beyond simple entity recognition, rendering them difficult to address with current methodologies.

% SP Drawbacks
\textbf{Resource scarcity for semantic parsing}. SP-based approaches require extensive annotated datasets for training, which are resource-intensive to create. This requirement severely limits the scalability of these methods and contributes to the "black box" nature of the reasoning process, which lacks transparency and interpretability.

\textbf{Underutilization of large language models (LLMs)}. Despite the demonstrated capabilities of LLMs in reasoning and few-shot learning, existing KBQA approaches have not fully leveraged these strengths. Most current methods either use LLMs as classifiers to identify predicates \citep{Gu-Yu-ACL-2023-Pangu, Jiashuo-Sun-ICLR-2024-Think-on-Graph} or as mimics to generate possible logical forms or questions \citep{Li-Tianle-ACL-2023-KB-BINDER, Li-Zhenyu-arXiv-2023-FlexKBQA}. There remains a significant opportunity to more effectively leverage LLMs in a few-shot setting to enhance the accuracy and interpretability of KBQA systems.

To address these challenges, we introduce the Interactive-KBQA framework, which combines the reasoning power of LLMs with three tools to interact with KBs. By conceptualizing the LLM as an agent and the KB as the environment, Interactive-KBQA facilitates an iterative, dialogue-based problem-solving process. Figure \ref{fig:model_overview} illustrates the overall process.
The process adheres to the thought-action paradigm. Given a complex query, the LLM is required to think and provide an action to interact with the KB through a set of specific tools. These tools return the execution results as observations. Specifically, we designed a set of tools that support heterogeneous databases (i.e., Freebase, Wikidata, and a Movie KB) with a unified interaction logic. 
We categorized complex questions and provided only two annotated exemplars with complete interactive processes for each type, serving as in-context learning demonstrations to prompt the LLM to complete the task. 
Furthermore, the method introduced in this study allowed for manual intervention.
Consequently, we manually annotated a small dataset with a detailed reasoning process, thereby creating a low-resource dataset. 
Finally, we fine-tuned open-source LLMs on this dataset. The experiments conducted have shown that this method is effective in low-resource contexts.
We have released this high-quality dataset to further contribute to NLP research in the community.

The main contributions of this work are summarized as follows: 
\begin{itemize}[noitemsep] 
    \item Propose the Interactive-KBQA, a novel framework that harnesses the reasoning capabilities of LLMs for semantic parsing, enabling multi-turn interactions with KBs. 
    \item Design a unified SPARQL-based toolset and interaction logic that efficiently address a wide array of complex queries. 
    \item Conduct extensive experiments to demonstrate that our method achieves remarkable performance with few exemplars. 
    \item Release a human-annotated KBQA dataset with step-wise reasoning processes, serving as a low-resource dataset.
\end{itemize}
\section{Related Work}

Knowledge base question answering (KBQA) methods can be broadly classified into two distinct categories: information retrieval (IR)-based methods and semantic parsing (SP)-based methods. These approaches address the challenge of effectively mapping natural language (NL) queries to the structured formats of knowledge bases (KBs).

% IR-based methods
\textbf{IR-based methods} aim to extract a question-specific subgraph from the KB and employ ranking algorithms to select the top entities or directly generate answers using text decoders \citep{Lan-Yunshi-TKDE-2023-ComplexKBQA-Survey}. To bridge the gap between unstructured texts and structured KBs, \citet{Dong-Guanting-CIKM-2023-Bridging-the-KB-Text-Gap} proposed a Structured Knowledge-aware Pre-training method to enhance complex subgraph representation learning. Emphasizing the accuracy of subgraph retrieval, \citet{Zhang-Jing-ACL-2022-Subgraph-Retrieval-Enhanced-Model} developed a trainable, decoupled subgraph retriever that boosts the performance of subgraph-oriented KBQA models. Further, \citet{Jinhao-Jiang-ICLR-2023-UniKGQA} introduced UniKGQA, a unified framework that integrates retrieval and reasoning across architectures and learning parameters.
% Drawbacks
However, IR-based methods typically rely on identifying entities within a query as an initial step for retrieving the relevant subgraph. This reliance poses a notable limitation, especially when handling complex queries.

% Semantic Parsing-based
\textbf{SP-based methods} parse questions into executable logical forms and perform queries against the KB to retrieve answers. Initial works translated questions into intermediate logical forms before execution. \citet{Yih-Wen-tau-ACL-2015-STAGG} defined a query graph that resembles subgraphs of the KB and can be directly mapped to a logical form. \citet{Hu-2018-TKDE-gAnswer, Lan-Yunshi-ACL-2020-Query-Graph-Generation} extended query graphs to include aggregation operators, and \citet{Chen-Yongrui-IJCAI-2021-Formal-Query-Building} enhanced candidate query generation by leveraging predictions of query structure.

% seq 2 seq
In recent years, generative models have increasingly been used to directly generate logical forms. \citet{Chen-Shuang-ACL-2021-ReTraCk} designed a retriever to fetch relevant KB items and utilized a Grammar-based Decoder based on LSTM for generating S-expressions \citep{Gu-Yu-WWW-2021-GrailQA}. Meanwhile, \citet{Das-Rajarshi-EMNLP-2021-Case-based-Reasoning} employed case-based reasoning for the KBQA task, using Big Bird \citep{Zaheer-Manzil-NIPS-2020-BigBird} to generate logical forms by retrieving relevant cases. Notably, works such as \citet{Ye-Xi-ACL-2022-RNG-KBQA}, \citet{Shu-Yiheng-EMNLP-2022-TIARA}, \citet{Zhang-Lingxi-ACL-2023-FC-KBQA}, and \citet{Yu-Donghan-ICLR-2023-DecAF} involved initially retrieving basic elements from the KB, such as entities, relations, subgraphs, or texts as supplementary information, before directly generating logical forms using the T5 model \citep{Raffel-JMLR-2020-T5}.
% Drawbacks
However, this approach requires substantial amounts of training data and suffers from a lack of transparency in the reasoning process.

% ----- LLM -----
\textbf{Large language models (LLMs) for KBQA} leverage the potent few-shot learning capabilities inherent in LLMs. Recent studies have shown that LLMs can significantly enhance reasoning over KBs. By capitalizing on the powerful few-shot learning capabilities, these methods have demonstrated marked improvements in both the accuracy and efficiency of information retrieval from KBs.

% icl
In the context of few-shot scenarios, \citet{Gu-Yu-ACL-2023-Pangu} proposed that LLMs should prioritize evaluating the plausibility of agent plans over directly generating answers. Concurrently, \citet{Li-Tianle-ACL-2023-KB-BINDER} advocated for the generation of logical forms as initial drafts, which are subsequently refined into executable queries using the KB.
% add (Reviewer mHdD 1)
% agent-env
From the agent-environment perspective, \citet{Jiang-EMNLP-2023-StructGPT} developed two specialized interfaces for accessing the KB, while \citet{Gu-Yu-arXiv-2024-Middleware} and \citet{Liu-Xiao-ICLR-2023-AgentBench} designed seven tools to facilitate this process. \citet{Jiashuo-Sun-ICLR-2024-Think-on-Graph} introduced a novel approach that enables LLMs to iteratively employ beam search reasoning on a KB.
% other
Furthermore, \citet{Zong-Chang-arXiv-2024-Triad} proposed assigning three distinct roles to the agent for addressing different KBQA subtasks. \citet{Jiang-Jinhao-arXiv-2024-KG-Agent} constructed an instruction dataset based on existing KBQA datasets.
% Drawbacks
However, despite employing LLMs, these methods lack the exploration of more complex questions.
\section{Approach}

\begin{figure*}[htbp]
  \centering
  \includegraphics[width=1\textwidth,page=2]{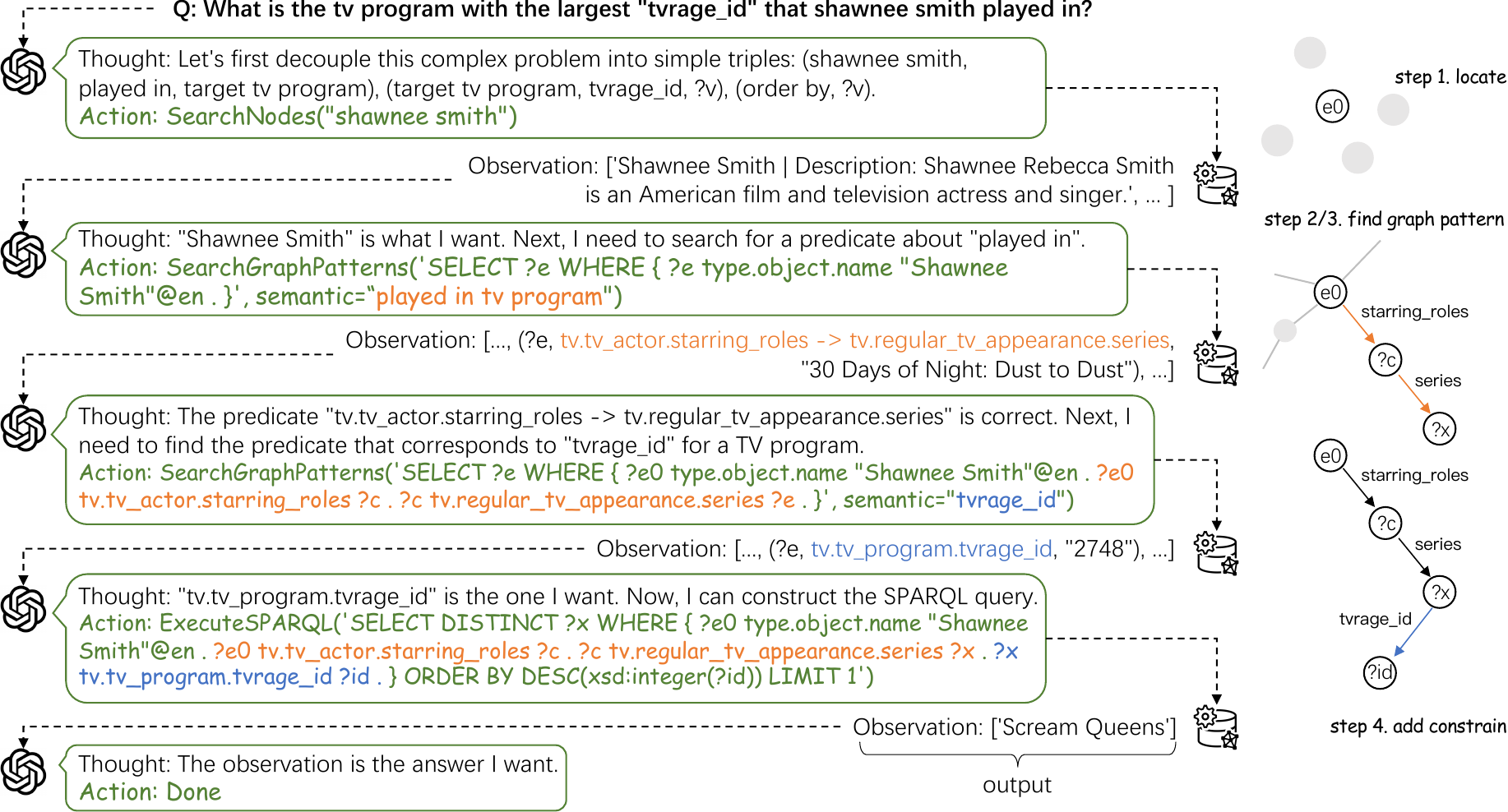}
  \caption{An example of the interactive process. \protect\footnotemark}
  \label{fig:dialog_overview}
\end{figure*}
\footnotetext{For brevity, we omit the prefix: \texttt{PREFIX ns: <http://rdf.freebase.com/ns/>}.}

% 3.1
\subsection{Problem Formulation}
This study investigates a semantic parsing (SP) method for knowledge base question answering (KBQA). A knowledge base (KB) is formally represented as $K \in E \times R \times (E \cup L \cup C)$, where $E$ denotes the set of entities, $R$ signifies the set of relations between entities, $C$ represents the set of classes, and $L$ includes the literal values.
Given a question $Q$ and a knowledge base $\mathcal{K}$, our goal is to generate an executable SPARQL expression $S$ that aligns with the question. 
Thus, the task can be formalized as $p(S|Q, \mathcal{K})$.

% 3.2
\subsection{Overview}

Recent advancements in large language models (LLMs) have demonstrated remarkable capabilities in few-shot learning and reasoning. 
However, fully leveraging LLMs to tackle complex KBQA challenges remains an elusive goal. 
To address this gap, we introduce Interactive-KBQA, an interactive method for KBQA that conceptualizes the LLM as an agent and the KB as an environment. 
This approach enables semantic parsing and SPARQL generation through dialogic interactions. Specifically, we design a unified interaction logic using three generic tools capable of supporting various complex types of questions (e.g., type constraints, count-based queries) across multiple databases (e.g., Freebase, Wikidata, Movie KB).
Figure \ref{fig:dialog_overview} presents an illustrative example of the interactive process.

% 3.3
\subsection{Tools for Knowledge Base}

The SP-based KBQA method necessitates the identification of elements and the appropriate graph patterns. Hence, it is crucial to design tools at an atomic level to ensure their universality. In line with this principle, we introduce the following three tools.

\textbf{SearchNodes(name)}: This function searches for nodes in the KB using the given surface name, \texttt{name}. Its primary purpose is entity linking (EL). Consequently, it not only returns the formal name of a node but also provides distinguishing features of the entity, such as its description and type. Importantly, this tool avoids dataset-specific EL techniques in favor of a generic retrieval approach. 

\textbf{SearchGraphPatterns(sparql, semantic)}: This function aims to identify and rank graph predicates within the KB that are essential, guided by the \texttt{semantic} parameter. The function requires the input \texttt{sparql}, which should consist of a SPARQL query beginning with "\texttt{SELECT ?e WHERE }". Following this, it performs a query on the one-hop subgraph centered around the entity ?e. This includes both incoming and outgoing edges. Subsequently, it ranks the retrieved triples based on the semantic relevance to the \texttt{semantic} parameter and the predicates in triples. Ultimately, the tool returns the top K triples.
% 注意 Solutions for Complex Questions 小节也提到了CVT优化
This tool is designed to precisely identify subgraphs while discarding irrelevant information, thereby optimizing the context window's use in LLMs. It supports flexible operations and is specifically optimized for Compound Value Type\footnote{A Compound Value Type is a Type within Freebase designed to represent data where each entry is composed of multiple fields.} (CVT) in Freebase by flattening a CVT node to multiple single-hop relationships. For example, to find movies featuring Tom Hanks in Freebase, the usage would be: \texttt{SearchGraphPatterns('SELECT ?e WHERE \{?e type.object.name "Tom Hanks"@en .\}', semantic="play in film")}, and return: \texttt{[(?e, film.actor.film -> film.performance.film, "Nothing in Common"), ...]}.

\textbf{ExecuteSPARQL(sparql)}: This tool allows for the direct execution of arbitrary SPARQL queries, ensuring unparalleled flexibility.

Implementation details are provided in Appendix \ref{app_sec:imp_details}, and additional usage instructions are described in Appendix \ref{app_sec:instruction_text}.

% 3.4
\subsection{Interactive Process}

Given a question $Q$, we first construct a prompt text:
\begin{equation}
    \text{Prompt} = \{\text{Inst}, E, Q\}
\end{equation}
where the instruction text, $\text{Inst}$, consists of tool descriptions, tool usage, and the format of interaction. $\text{Inst}$ is database-specific and pre-written. $E$ denotes a set of exemplars, and for each type of question, we manually annotate two complete examples in an interactive format.

In each turn $T$, we let the LLM generate an action based on the $\text{Prompt}$ and the history $H$ of the interaction. Specifically, this procedure follows:
\begin{equation}
    \label{eq:interactive_process}
    a_T = \text{LLM}(\{\text{Prompt}, H\})
\end{equation}
\begin{equation}
    H = \{c_0,a_0,o_0, ...,c_{T-1}, a_{T-1}, o_{T-1}\}
\end{equation}

where $c$ denotes the intermediate thought process, an action $a$ belongs to the set \{SearchNodes, SearchGraphPatterns, ExecuteSPARQL, Done\}, and the observation $o$ is determined by executing an action, which is defined as $o_T = \text{Tool}(a_T)$.

We have devised a general thought-action paradigm for KBQA. 
\textbf{Thought}: Given $Q$, our initial thought $c_0$ is to deconstruct it into sub-queries reminiscent of triple forms; for example, the aforementioned example can be decomposed into \texttt{(Tom Hanks, act in, target movie)}. $c_0$ is not rigidly defined but adopts a free-form approach, leveraging the semantics of the question to facilitate understanding by both LLMs and humans.
Except for the first round, the LLM is required to generate a thought $c$ that clearly articulates its reasoning process based on observations. This approach is intended to render the decision-making process explainable.
\textbf{Action}: At each turn $T$, the LLM must generate an action $a_T$ that concludes the current round, using Python syntax. The tool then parses, executes, and returns the results, serving as the observation\footnote{In case of parsing errors, the tool will return a specific error message.}.
The LLM decides whether to end the dialogue based on the observation. If $a_T = \text{Done}$, we output the final observation $o_T$ as the answer.

To minimize inference costs, we train a question type classifier on low-resource data and select $E$ based on the predicted type, as further discussed in Appendix \ref{app_sec:imp_details}.

% 3.5
\subsection{Solutions for Complex Questions}

Interactive-KBQA involves guiding LLMs through reasoning by annotating an interactive inference process. For different types of complex questions, it is crucial to identify patterns, design interaction modes, and label high-quality examples. This section outlines several representative types of complex questions to elucidate the design rationale behind this paper.

For \textbf{multi-hop} questions, our focus at each step is on specific predicates rather than concrete entities. For instance, as depicted in Figure \ref{fig:dialog_overview}, there is no need to locate specific television programs; expressing the graph pattern within SPARQL is sufficient to enable the tool to handle the ranking part of the task.
In the case of \textbf{CVT structures} in Freebase, we explicitly describe the reasoning process when encountered. Moreover, we break down the star-shaped CVT structure into multiple one-hop relations and treat them accordingly. 
% 和 SearchGraphPatterns 小节一致
For instance, the semantics of the sentence ``Tom Hanks plays the role of `David Basner' in the film `Nothing in Common' '' can be represented by two triples: (``Tom Hanks'', film.actor.film $\rightarrow$ film.performance.film, ``Nothing in Common'') and (``Tom Hanks'', film.actor.film $\rightarrow$ film.performance.character, ``David Basner'').
When querying the \textbf{qualifier structure} in Wikidata, which modifies predicates, we devise a specialized SPARQL query pattern and provide a detailed thought process as well.
Additional examples are provided in the Appendix \ref{app_sec:case_study}.

% 3.6
\subsection{Human-Machine Collaborative Annotation}

% motivation 
In realistic scenarios, including examples of all question types within the context is impractical for two primary reasons. First, the associated costs are substantial. Second, as discussed by \citet{Su-Jianlin-arXiv-2023-RoPE, Zhu-Dawei-arXiv-2023-PoSE}, LLMs experience a notable decline in performance when input tokens exceed certain limits.
Additionally, \citet{Lightman-Hunter-arXiv-2023-ProcessSupervision} pointed out that process supervision can enhance a model's generalization capabilities.
Consequently, we propose a human-machine collaborative data annotation method. Enabled by the Interactive-KBQA method introduced in this paper, annotating the reasoning process has become more straightforward.

% methodology
Specifically, when annotators determine that the action $a_T$ is unreasonable, it is manually corrected to $a'_T$. This adjustment is then incorporated into the message to generate the action $a_{T+1}$. Formally,

\begin{equation*}
\resizebox{0.48\textwidth}{!}{$
a_{T+1}=\text{LLM}(\{\text{Prompt}, \{c_0,a_0,o_0, ...,c_{T-1}, a'_{T}, o'_{T}\} \})
$}
\end{equation*}

We set a breakpoint after each round where the LLM generates thoughts and actions, allowing human evaluators to review and decide on their acceptance. Once accepted, the process continues. If rejected, the annotator revises the generated thoughts and actions before proceeding. For each question type, we randomly sampled 50 data points from the training set and manually annotated them to provide low-resource data.

% criteria
The core principle of our methodology is to emulate the human data annotation process. 
Annotators intervene in specific scenarios where the model exhibits hallucinations (such as generating predicates not present in the observation), inconsistencies between Thought and Action, or deviations from the correct answer trajectory (for instance, cases where two consecutive rounds are incorrect).
\section{Experiment}

We examine Interactive-KBQA across a variety of complex question types and diverse databases (DBs).

% 4.1
\subsection{Dataset \& Preprocessing}

\begin{table}
\scalebox{0.75}{
    \begin{tabular}{cccc}
    \hline
    \textbf{Dataset} & \textbf{\#Type} & \textbf{\begin{tabular}[c]{@{}c@{}}\#Anno\\ (Train/Test)\end{tabular}} & \textbf{\begin{tabular}[c]{@{}c@{}}\#Raw\\ (Train/Dev/Test)\end{tabular}} \\ \hline
    WebQSP & 2 & 100 / 300 & 3,098 / - / 1,639 \\
    CWQ & 4 & 200 / 600 & 27,639 / 3,519 / 3,531 \\
    KQA Pro & 9 & 450 / 900 & 94,376 / 11,797 / 11,797 \\
    MetaQA & 3 & 150 / 900 & 329,282 / 39,138 / 30,903 \\ \hline
    \end{tabular}
}
\caption{Statistics of the datasets.}
\label{tab:data_statistics}
\end{table}

\textbf{WebQuestionsSP} (WebQSP) \citep{Yih-Wen-tau-ACL-2016-WebQSP} and \textbf{ComplexWebQuestions 1.1} (CWQ) \citep{Talmor-Alon-NAACL-2018-ComplexWebQuestions-CWQ} are extensively used in KBQA research. These datasets comprise natural language questions paired with their corresponding SPARQL queries based on Freebase. 
Following \citet{Chen-Zi-Yuan-NAACL-2019-UHop}, we classify WebQSP questions into 1-hop and 2-hop categories based on the length of the inferential relation chain, i.e., the path from the topic entity to the answer node. 
CWQ extends WebQSP by incorporating four types of complex questions: Conjunction (Conj), Composition (Compo), Comparative (Compa), and Superlative (Super).

\textbf{KQA Pro} \citep{Cao-Shulin-ACL-2022-KQAPro} is a large-scale dataset designed for complex question answering over a dense set of Wikidata entries. It features nine types of complex questions, including Count (Ct), Query Attribute (QA), Query Attribute Qualifier (QAQ), Query Name (QN), Query Relation (QR), Query Relation Qualifier (QRQ), Select Among (SA), Select Between (SB), and Verify (Vf).

\textbf{MetaQA} \citep{Zhang-Yuyu-AAAI-2018-MetaQA} is built upon a Movie Knowledge Base (KB) and includes three sets of question-answer pairs: 1-hop, 2-hop, and 3-hop. We converted the Movie KB into RDF triples to facilitate querying via SPARQL.

% 采样
Due to the prohibitive costs associated with utilizing OpenAI, we employed a uniform sampling method across each type to construct a smaller test dataset. For each DB, we sampled 900 instances, distributing these as follows: 150 instances each for WebQSP and CWQ, 100 instances for KQA Pro, and 300 instances for MetaQA.
The statistical details of the test dataset and our annotated (Anno) dataset are presented in Table~\ref{tab:data_statistics}.
Our sampled dataset presents a balanced composition, enhancing the assessment of models' capabilities in handling complex questions.

% 4.2
\subsection{Baselines}

To comprehensively evaluate our approach, we have selected a range of prior state-of-the-art (SOTA) baseline models.

\textbf{Fine-tuning (FT) on full data}. We selected semantic parsing (SP)-based methods as baselines. DeCAF \citep{Yu-Donghan-ICLR-2023-DecAF} generates both logical forms and direct answers to provide the final outcomes. For KQA Pro, we opted for the BART-SPARQL \citep{Cao-Shulin-ACL-2022-KQAPro}. For MetaQA, we choose Edge-aware \citep{Zhang-Yanan-Edge-aware-GNN-Neurosci-2022}.

\textbf{Prompting methods}. We chose KB-BINDER \citep{Li-Tianle-ACL-2023-KB-BINDER} which  leverage large language models (LLMs) in a few-shot setting. 
Additionally, we compared the efficacy of Chain-of-Thought prompting (CoT prompt) \citep{Wei-Jason-NeurIPS-2022-ChainOfThought} alongside the technique of Self-Consistency (SC) \citep{Wang-Xuezhi-ICLR-2023-SelfConsistency}, further enriching our analysis.
It's important to note that StructGPT \citep{Jiang-EMNLP-2023-StructGPT} and ToG \citep{Jiashuo-Sun-ICLR-2024-Think-on-Graph} operate under the assumption that golden entities have been provided. Consequently, the comparison of these methodologies is presented in Section \ref{sec:el_and_golden}.

\textbf{FT on low-resource data}. We fine-tuned open-source LLMs (open-LLMs) to directly generate SPARQL queries through supervised fine-tuning (SFT) manner. We reimplemented a prior SOTA approach for each dataset to serve as a comparison.

% 4.3
\subsection{Evaluation Metrics}
SP-based methods generate logical forms, which consequently produce answers in an unordered manner. These should be evaluated using the F1 score\footnote{We use the average of the F1 scores across all instances.}.
Additionally, we also report the Random Hits@1 (RHits@1)\footnote{Randomly selecting an answer for each question 100 times and calculating the average Hits@1.} \citep{Shu-Yiheng-EMNLP-2022-TIARA}, and the Exact Match (EM) metric \citep{Talmor-Alon-NAACL-2018-ComplexWebQuestions-CWQ} for reference. For KQA Pro, we report the accuracy, defined as the condition where the two sets match exactly.

% 4.4
\subsection{Results}

\begin{table*}[htp]
\centering
\scalebox{0.72}{
    \begin{tabular}{clcccccccccc}
    \hline
    \multicolumn{2}{c}{\multirow{2}{*}{\textbf{Method}}} & \multicolumn{4}{c}{\textbf{WebQSP}} & \multicolumn{6}{c}{\textbf{CWQ}} \\ \cline{3-12} 
    \multicolumn{2}{c}{} & \textbf{1-hop} & \textbf{2-hop} & \textbf{Overall} & \textbf{RHits@1} & \textbf{Conj} & \textbf{Compo} & \textbf{Compa} & \textbf{Super} & \textbf{Overall} & \textbf{EM} \\ \hline
    Prior FT SOTA & DeCAF $\dagger$ & 74.72 & 76.32 & 75.52 & 80.28 & 69.19 & 53.54 & 18.04 & 28.00 & 42.19 & 50.83 \\
    Prompting SOTA & KB-BINDER$\sharp$ & \multicolumn{2}{c}{-} & 74.40 & - & \multicolumn{6}{c}{-} \\ \hline
    \multirow{4}{*}{\begin{tabular}[c]{@{}c@{}}Prompting\\ w/GPT-4 Turbo\end{tabular}} & IO & 28.54 & 50.05 & 39.29 & 45.51 & 47.54 & 29.71 & 33.66 & 24.67 & 33.89 & 45.67 \\
     & CoT & 27.85 & 51.55 & 39.70 & 47.52 & 44.12 & 26.30 & 34.39 & 30.00 & 33.70 & 43.67 \\
     & CoT+SC & 26.66 & 51.35 & 39.01 & 47.08 & \textbf{50.65} & 28.78 & 36.98 & 29.78 & 36.55 & 47.50 \\
     & Ours & \textbf{69.99} & \textbf{72.41} & \textbf{71.20} & \textbf{72.47} & 47.44 & \textbf{59.00} & 47.89 & 41.96 & \textbf{49.07} & \textbf{59.17} \\ \hline
    Reimplement & DeCAF & 24.56 & 27.55 & 27.55 & 39.33 & 32.19 & 10.16 & 11.63 & 6.00 & 15.00 & 19.50 \\ \hline
    \multirow{3}{*}{\begin{tabular}[c]{@{}c@{}}Fine-tuning\\ w/open-LLM\end{tabular}} & SFT-SPARQL (7B) & 34.39 & 33.80 & 34.09 & 35.68 & 12.39 & 20.44 & 41.10 & 38.44 & 28.10 & 30.00 \\
     & Ours (7B) & 42.02 & 45.03 & 43.57 & 45.09 & 31.90 & 30.70 & 50.98 & 46.03 & 39.90 & 44.00 \\
     & Ours (13B) & 55.81 & 53.92 & 54.86 & 56.25 & 30.47 & 34.51 & \textbf{55.98} & \textbf{49.06} & 42.50 & 45.67 \\ \hline \hline
    \multicolumn{2}{c}{\multirow{2}{*}{\textbf{Method}}} & \multicolumn{10}{c}{\textbf{KQA Pro}} \\ \cline{3-12} 
    \multicolumn{2}{c}{} & \textbf{Ct} & \textbf{QA} & \textbf{QAQ} & \textbf{QN} & \textbf{QR} & \textbf{QRQ} & \textbf{SA} & \textbf{SB} & \textbf{Vf} & \textbf{Overall} \\ \hline
    Prior FT SOTA & BART-SPARQL $\dagger$ & 89 & 92 & 87 & 77 & 95 & 81 & 96 & 94 & 86 & 88.56 \\ \hline
    \multirow{4}{*}{\begin{tabular}[c]{@{}c@{}}Prompting\\ w/GPT-4 Turbo\end{tabular}} & IO & 27 & 23 & 36 & 40 & 25 & 50 & 11 & 69 & 73 & 39.33 \\
     & CoT & 22 & 26 & 35 & 34 & 18 & 46 & 21 & 79 & 77 & 39.78 \\
     & CoT+SC & 25 & 28 & 33 & 38 & 22 & 51 & 19 & 86 & 75 & 41.89 \\
     & Ours & \textbf{74} & \textbf{83} & 64 & \textbf{73} & 73 & \textbf{59} & \textbf{80} & 61 & \textbf{80} & \textbf{71.89} \\ \hline
    Reimplement & BART-SPARQL & 37 & 44 & 37 & 36 & 67 & 33 & 49 & 78 & 58 & 48.78 \\ \hline
    \multirow{3}{*}{\begin{tabular}[c]{@{}c@{}}Fine-tuning\\ w/open-LLM\end{tabular}} & SFT-SPARQL (7B) & 52 & 51 & 52 & 47 & 69 & 37 & 60 & \textbf{85} & 67 & 57.78 \\
     & Ours (7B) & 53 & 58 & \textbf{69} & 48 & 75 & 48 & 75 & 84 & 70 & 64.40 \\
     & Ours (13B) & 63 & 65 & \textbf{55} & 49 & \textbf{76} & 52 & 68 & 75 & 62 & 62.78 \\ \hline
    \end{tabular}
}
\caption{Results on WebQSP and CWQ. Results tagged with $\dagger$ denote data from original prediction files, but evaluated on consistent test data. Results with $\sharp$ are reprinted from \protect\citep{Li-Tianle-ACL-2023-KB-BINDER}.}
\label{tab:main_res_fb_kqa}
\end{table*}

\begin{table}[htp]
\centering
\scalebox{0.72}{
    \begin{tabular}{cccccc}
    \hline
    \multirow{2}{*}{\textbf{Method}} & \multicolumn{5}{c}{\textbf{MetaQA}} \\ \cline{2-6} 
     & \textbf{1-hop} & \textbf{2-hop} & \textbf{3-hop} & \textbf{Overall} & \textbf{RHits@1} \\ \hline
    Edge-aware $\sharp$ & 98.50 & 93.70 & 91.00 & 94.40 & 96.77 \\
    KB-BINDER $\ddagger$ & 82.15 & 91.26 & 99.66 & 91.02 & 86.52 \\ \hline
    Ours w/GPT-4 & \textbf{96.75} & \textbf{98.47} & \textbf{93.55} & \textbf{96.25} & \textbf{95.97} \\
    Ours w/SFT (7B) & 93.89 & 85.99 & 95.61 & 91.83 & 91.41 \\ \hline
    \end{tabular}
}
\caption{Results on MetaQA. The symbol $\sharp$ denotes results reprinted from \protect\citet{Zhang-Yanan-Edge-aware-GNN-Neurosci-2022}, while $\ddagger$ indicates we reimplemented the results.}
\label{tab:main_res_metaqa}
\end{table}

\paragraph{Prompting with GPT-4 Turbo}
Table \ref{tab:main_res_fb_kqa} and Table \ref{tab:main_res_metaqa} present comprehensive comparisons.
As indicated in the table, compared to methods that utilize full data, our approach is at a natural disadvantage on the WebQSP and KQA Pro datasets due to the significant difference in the magnitude of training data (\textasciitilde3K/\textasciitilde33K vs. 4-shot/2-shot, respectively).
However, on CWQ and MetaQA datasets, our approach overall outperforms baselines. 
Furthermore, in the cases of comparative and superlative question types, our approach achieves improvements of 29.85\% and 13.96\%, respectively. 
This can be attributed to the fact that within the original dataset, these two types of questions each constitute merely 5\% (see Table \ref{app_tab:distribution_qtype_cwq}).
Unlike previous methods, our approach is not constrained by data distribution, rendering it more robust.
% add ???
For conjunction questions in CWQ, CoT+SC and Decaf outperform our method due to redundant constraints inherent in these queries. Such questions require multiple entity constraints (e.g., "What country in the Mediterranean has Zonguldak Province?") yet often only one is needed to pinpoint the answer ("Zonguldak Province" leads to "Turkey", rendering "country in the Mediterranean" extraneous). Our approach, which focuses on identifying each predicate, complicates the response process.

\paragraph{FT with open-LLMs}
The results show that our method outperforms all baselines in low-resource scenarios. Specifically, for evaluations involving two question types from CWQ and two types from KQA Pro, the fine-tuned model exceeded the performance of GPT-4 Turbo.

% 4.5
% entity linking
\subsection{Impact of Entity Linking (EL)} \label{sec:el_and_golden}

% add ELQ (Reviewer F8iA 4)
\begin{table}
\centering
\scalebox{0.96}{
    \begin{tabular}{lcccc}
        \hline
        \textbf{Dataset} & \textbf{Precision} & \textbf{Recall} & \textbf{F1} & \textbf{MCR} \\ \hline
        WebQSP & 91.50  & 74.69  & 80.00  & 67.42  \\
        \multicolumn{1}{r}{w/ELQ} & 93.67  & 38.69  & 41.30  &    \\
        CWQ & 87.92  & 70.67  & 76.06  & 76.64  \\
        \multicolumn{1}{r}{w/ELQ} & 94.36  & 41.61  & 43.81  &    \\
        KQA Pro & 80.91  & 75.82  & 75.35  & 80.80  \\
        MetaQA & 97.33  & 95.89  & 95.89  & 100.00  \\ \hline
    \end{tabular}
}
\caption{Results of entity linking.}
\label{tab:res_el}
\end{table}

\begin{table}
\centering
\scalebox{0.85}{
    \begin{tabular}{lcccc}
    \hline
    \multicolumn{1}{c}{\multirow{2}{*}{\textbf{Methods}}} & \multicolumn{2}{c}{\textbf{WebQSP}} & \multicolumn{2}{c}{\textbf{CWQ}} \\ \cline{2-5} 
    \multicolumn{1}{c}{} & \textbf{Overall} & \textbf{RHits@1} & \textbf{Overall} & \textbf{EM} \\ \hline
    StructGPT $\dagger$ & 44.26 & -\footnotemark & \multicolumn{2}{c}{-} \\
    ToG $\ddagger$ & 36.40 & 44.80 & 31.77 & 41.94 \\ \hline
    Ours & 71.20 & 72.47 & 49.07 & 59.17 \\
    Ours w/golden & 78.64 & 79.25 & 56.74 & 66.50 \\
    Gain & 7.44 & 6.77 & 7.67 & 7.33 \\ \hline
    \end{tabular}
}
\caption{Results on WebQSP and CWQ with golden entities. $\dagger$ indicates original predictions evaluated on a consistent dataset; $\ddagger$ denotes our reimplementation.}
\label{tab:res_golden_entity}
\end{table}
\footnotetext{The output of StructGPT consists of a string that contains answers, making it unsuitable for evaluation with RHits@1.}

The EL performance of our methods\footnote{We find the entity surface name rather than mid.} (with GPT-4 Turbo) is delineated in Table \ref{tab:res_el}. 
% add ELQ (Reviewer F8iA 4)
The table reports a comparison of the results of our method with those of the widely recognized entity linking tool, ELQ \citep{Li-Belinda-EMNLP-2020-ELQ}, on the WebQSP and CWQ datasets.
Moreover, we introduce the Mention Cover Rate (MCR) to quantify the difficulty of EL.
MCR is defined as the rate at which the golden entity names appear within the questions. 
As shown in our analysis, KQA Pro and MetaQA exhibit higher MCR. 
It is noted that some entity constrains are redundant in KQA Pro\footnote{For instance, the answer of "What is the connection between Steve Jordan (the one whose position is tight end) and Phoenix (the one that is the twinned administrative body of Chengdu)?" is equivalent to that of "What is the connection between Steve Jordan and Phoenix?".}.
Therefore, despite the F1 score being 75.35\%, this is not a bottleneck affecting the performance of KQA Pro. 
Based on these observations, we conducted experiments on the WebQSP and CWQ datasets under the conditions of given golden entities. 
The results, as presented in Table \ref{tab:res_golden_entity}, reveal that our approach significantly outperforms the previous SOTA across both datasets.

\subsection{Ablation Study}
We perform various ablation studies to understand the importance of different factors in Interactive-KBQA. 

\paragraph{Impact of Exemplars}\label{sec:ablation_shot}

\begin{table}
\centering
\scalebox{1}{
    \begin{tabular}{lccc}
    \hline
    \textbf{Dataset} & \textbf{Precision} & \textbf{Recall} & \textbf{F1} \\ \hline
    CWQ & 92.76 & 92.83 & 92.79 \\
    KQA Pro & 92.73 & 92.44 & 92.42 \\ \hline
    \end{tabular}
}
\caption{The performances of question type classifiers.}
\label{tab:res_bert_cls}
\end{table}

\begin{table}[htp]
\centering
\scalebox{0.9}{
    \begin{tabular}{rllll}
    \hline
    \multicolumn{1}{c}{\multirow{2}{*}{\textbf{Setting}}} & \multicolumn{2}{c}{\textbf{CWQ}} & \multicolumn{2}{c}{\textbf{KQA Pro}} \\ \cline{2-5} 
    \multicolumn{1}{c}{} & \multicolumn{1}{c}{\textbf{F1}} & \multicolumn{1}{c}{\textbf{AP}} & \multicolumn{1}{c}{\textbf{Acc}} & \multicolumn{1}{c}{\textbf{AP}} \\ \hline
    \multicolumn{1}{l}{Ours(cls+2-shot)} & 54.69 & \$0.50 & 67.78 & \$0.38 \\
    w/4-shot $\alpha$ & 57.19 & \$0.70 & 46.67 & \$0.50 \\
    w/4-shot $\beta$ & \multicolumn{2}{c}{-} & 52.22 & \$0.55 \\
    zero-shot & 51.83 & \$0.37 & 25.25 & \$0.30 \\ \hline
    \end{tabular}
}
\caption{The impact of exemplar number and the average price (AP). The KQA Pro configuration $\alpha$ includes QN, QR, QRQ, and Vf, Whereas $\beta$ comprises Ct, QAQ, SA, and SB.}
\label{tab:ablation_shot}
\end{table}

% add (Reviewer F8iA 1)
This section conducts an ablation study to examine the influence of exemplars in two representative settings: CWQ, which includes four types and can barely cover all question types within the prompt text, and KQA Pro, which includes nine types and cannot cover all types due to high inference costs.

The performance of classifiers (cls) trained with our annotated dataset is shown in Table \ref{tab:res_bert_cls}. 
Experiments were conducted in 0-shot and 4-shot scenarios on a randomly selected test subset of 100 entries.
For CWQ, one exemplar per question type was used, while for KQA Pro, only four types were sampled. 
In the zero-shot scenario, only the instruction text was used as a prompt. 
As Table \ref{tab:ablation_shot} shows, our method's performance improves with increased question type coverage in prompts. 
For CWQ 4-shots, we observed a 2.5 point performance increase, but the cost rose by 37.86\%, from \$0.5 to \$0.7. 
KQA Pro results suggest that accurate demonstrations improve performance and reduce costs. 
For a detailed discussion about the interaction rounds and costs, see Appendix \ref{app_sec:round_price}.

\paragraph{Impact of Backbone Model}

\begin{table}[htp]
\centering
\scalebox{1}{
    \begin{tabular}{rcc}
    \hline
    \multicolumn{1}{l}{\textbf{Model}} & \textbf{CWQ} & \textbf{KQA Pro} \\ \hline
    \multicolumn{1}{l}{OpenAI} & \multicolumn{1}{l}{} & \multicolumn{1}{l}{} \\
    GPT-4 Turbo & 49.07 & 71.89 \\
    GPT-3.5 Turbo & 13.42 & 47.92 \\ \hline
    \multicolumn{1}{l}{open-source LLM} &  &  \\
    Mistral-7B FT & 39.90 & 62.24 \\
    w/o FT & 4.76 & 20.41 \\
    Llama 2 7B FT \footnotemark & 30.42 & 66.33 \\
    Llama 2 13B FT & 42.50 & 62.78 \\ \hline
    \end{tabular}
}
\caption{Performance of different backbone models}
\label{tab:ablation_backbone_llm}
\end{table}
\footnotetext{The maximum context of Llama 2, which is 4,096 tokens, is insufficient for direct inference.}

As demonstrated in Table \ref{tab:ablation_backbone_llm}, GPT-4 significantly outperforms GPT-3.5\footnote{We use gpt-3.5-turbo-1106.} in terms of reasoning capabilities. 
We have also attempted to apply direct reasoning with Mistral 7B. It was observed that the untrained model significantly struggles with complex, multi-turn interactions.
This finding underscores the substantial improvements that fine-tuning provides.

% 4.7
\subsection{Error Analysis}

\begin{table}
\centering
\scalebox{1}{
    \begin{tabular}{lcc}
    \hline
    \textbf{Error Type} & \textbf{WCWQ} & \textbf{KQA Pro} \\ \hline
    Entity Linking & 18 & 7 \\
    Predicate Search & 6 & 0 \\
    Reasoning Error & 32 & 48 \\
    Format Compliance & 17 & 15 \\
    Hallucination & 19 & 21 \\
    Other & 8 & 9 \\ \hline
    \end{tabular}
}
\caption{Distribution of error types.}
\label{tab:error_analysis}
\end{table}

To systemically assess our method's deficiencies, we first amalgamate WebQSP+CWQ (WCWQ) and randomly select 100 error instances from each for manual inspection. The aggregated statistical findings are detailed in Table \ref{tab:error_analysis}.

% add (Reviewer mHdD 3)
\begin{itemize}[noitemsep, leftmargin=*]

\item \textbf{Entity Linking Error} refers to the failure to locate nodes using the SearchNodes tool.
The primary cause of this error is the LLM's failure to extract the correct entity mention from the question. For example, in the question "Most Anticipated Tour at the Young Hollywood Awards.", the entity should have been identified as "Young Hollywood Award for Most Anticipated Tour", but instead, the LLM mistakenly searched for "Young Hollywood Awards".
Besides, In CWQ and WebQSP, the entities in the returned results do not have descriptions, leading the LLM to consider the results incorrect and subsequently re-initiate the search.
\item \textbf{Predicate Search Error} denotes the failure of the SearchGraphPatterns tool to return the necessary information.
This error arises when vector search tools fail to return expected results, such as predicates indicating an organization's headquarters. For instance, the query "locate in" is unable to match the predicate "organization.organization.headquarters".
\item \textbf{Reasoning Error} means that, given the observations, the LLM fails to generate the appropriate SPARQL query.
This failure is largely due to an insufficient semantic understanding of the KG schema. Typically, the LLM often fails to understand the CVT structure (e.g. the direction of the predicate), resulting in incomplete or inaccurate SPARQL queries.
\item \textbf{Format Compliance Error} implies that the LLM does not use the tool in the required format.
Examples include incorrectly constructed SPARQL queries with improperly formatted time and numerical values, and the introduction of unrecognized additional parameters to the tool.
\item \textbf{Hallucination} includes generating elements that are inconsistent with the observations.
\item \textbf{Other Error} encompasses errors that cannot be categorized under the above types.

\end{itemize}

More details are presented in the form of case study in the Appendix \ref{app_sec:case_study}.
\section{Conclusion} Interactive-KBQA introduces a KBQA approach which utilizes an LLM as an agent for performing semantic parsing through multi-round interactions with a KB. Initially, we developed a unified tool and an interaction methodology tailored to various DB schemas. Subsequently, by categorizing complex questions and annotating a minimal set of exemplars, we employed a few-shot learning strategy that guides the LLM to incrementally generate SPARQL queries. Moreover, we introduced a low-resource dataset that demonstrates superior performance when fine-tuned with open-source LLMs.

\section*{Limitations}
The prompt learning-based approach heavily relies on the capabilities of LLMs, and in scenarios involving multiple rounds of dialogue, the cost of reasoning becomes significantly high. 
Additionally, it is impractical to adjust the output of LLM when invoking LLM APIs. 
Therefore, this paper proposes a collaborative human-machine annotation method to mitigate this issue.

\section*{Acknowledgments}
This work was supported by the National Key Research and Development Program of China under Grant No. 2023YFC3304404.
We would also like to thank to Prof. Zhiyuan Liu and Dr. Yujia Qin from THUNLP group for their invaluable suggestions.

% Bibliography entries for the entire Anthology, followed by custom entries
%\bibliography{anthology,custom}
% Custom bibliography entries only
\bibliography{references}

\begin{thebibliography}{48}
\expandafter\ifx\csname natexlab\endcsname\relax\def\natexlab#1{#1}\fi

\bibitem[{Bollacker et~al.(2008)Bollacker, Evans, Paritosh, Sturge, and Taylor}]{Bollacker-Kurt-SIGMOD-2008-Freebase}
Kurt Bollacker, Colin Evans, Praveen Paritosh, Tim Sturge, and Jamie Taylor. 2008.
\newblock \href {https://doi.org/10.1145/1376616.1376746} {Freebase: a collaboratively created graph database for structuring human knowledge}.
\newblock In \emph{Proceedings of the 2008 ACM SIGMOD International Conference on Management of Data}, SIGMOD '08, page 1247–1250, New York, NY, USA. Association for Computing Machinery.

\bibitem[{Cao et~al.(2022)Cao, Shi, Pan, Nie, Xiang, Hou, Li, He, and Zhang}]{Cao-Shulin-ACL-2022-KQAPro}
Shulin Cao, Jiaxin Shi, Liangming Pan, Lunyiu Nie, Yutong Xiang, Lei Hou, Juanzi Li, Bin He, and Hanwang Zhang. 2022.
\newblock \href {https://doi.org/10.18653/v1/2022.acl-long.422} {{KQA} pro: A dataset with explicit compositional programs for complex question answering over knowledge base}.
\newblock In \emph{Proceedings of the 60th Annual Meeting of the Association for Computational Linguistics (Volume 1: Long Papers)}, pages 6101--6119, Dublin, Ireland. Association for Computational Linguistics.

\bibitem[{Chen et~al.(2021{\natexlab{a}})Chen, Liu, Yu, Lin, Lou, and Jiang}]{Chen-Shuang-ACL-2021-ReTraCk}
Shuang Chen, Qian Liu, Zhiwei Yu, Chin-Yew Lin, Jian-Guang Lou, and Feng Jiang. 2021{\natexlab{a}}.
\newblock \href {https://doi.org/10.18653/v1/2021.acl-demo.39} {{R}e{T}ra{C}k: A flexible and efficient framework for knowledge base question answering}.
\newblock In \emph{Proceedings of the 59th Annual Meeting of the Association for Computational Linguistics and the 11th International Joint Conference on Natural Language Processing: System Demonstrations}, pages 325--336, Online. Association for Computational Linguistics.

\bibitem[{Chen(2023)}]{Chen-Wenhu-EACL-2023-LLMTableReasoners}
Wenhu Chen. 2023.
\newblock \href {https://doi.org/10.18653/v1/2023.findings-eacl.83} {Large language models are few(1)-shot table reasoners}.
\newblock In \emph{Findings of the Association for Computational Linguistics: EACL 2023}, pages 1120--1130, Dubrovnik, Croatia. Association for Computational Linguistics.

\bibitem[{Chen et~al.(2021{\natexlab{b}})Chen, Li, Hua, and Qi}]{Chen-Yongrui-IJCAI-2021-Formal-Query-Building}
Yongrui Chen, Huiying Li, Yuncheng Hua, and Guilin Qi. 2021{\natexlab{b}}.
\newblock Formal query building with query structure prediction for complex question answering over knowledge base.
\newblock In \emph{Proceedings of the Twenty-Ninth International Joint Conference on Artificial Intelligence}, IJCAI'20.

\bibitem[{Chen et~al.(2019)Chen, Chang, Chen, Nayak, and Ku}]{Chen-Zi-Yuan-NAACL-2019-UHop}
Zi-Yuan Chen, Chih-Hung Chang, Yi-Pei Chen, Jijnasa Nayak, and Lun-Wei Ku. 2019.
\newblock \href {https://doi.org/10.18653/v1/N19-1031} {{UH}op: An unrestricted-hop relation extraction framework for knowledge-based question answering}.
\newblock In \emph{Proceedings of the 2019 Conference of the North {A}merican Chapter of the Association for Computational Linguistics: Human Language Technologies, Volume 1 (Long and Short Papers)}, pages 345--356, Minneapolis, Minnesota. Association for Computational Linguistics.

\bibitem[{Das et~al.(2021)Das, Zaheer, Thai, Godbole, Perez, Lee, Tan, Polymenakos, and McCallum}]{Das-Rajarshi-EMNLP-2021-Case-based-Reasoning}
Rajarshi Das, Manzil Zaheer, Dung Thai, Ameya Godbole, Ethan Perez, Jay~Yoon Lee, Lizhen Tan, Lazaros Polymenakos, and Andrew McCallum. 2021.
\newblock \href {https://doi.org/10.18653/v1/2021.emnlp-main.755} {Case-based reasoning for natural language queries over knowledge bases}.
\newblock In \emph{Proceedings of the 2021 Conference on Empirical Methods in Natural Language Processing}, pages 9594--9611, Online and Punta Cana, Dominican Republic. Association for Computational Linguistics.

\bibitem[{Dong et~al.(2023)Dong, Li, Wang, Zhang, Xian, and Xu}]{Dong-Guanting-CIKM-2023-Bridging-the-KB-Text-Gap}
Guanting Dong, Rumei Li, Sirui Wang, Yupeng Zhang, Yunsen Xian, and Weiran Xu. 2023.
\newblock \href {https://doi.org/10.1145/3583780.3615150} {Bridging the kb-text gap: Leveraging structured knowledge-aware pre-training for kbqa}.
\newblock In \emph{Proceedings of the 32nd ACM International Conference on Information and Knowledge Management}, CIKM '23, page 3854–3859, New York, NY, USA. Association for Computing Machinery.

\bibitem[{Gu et~al.(2023)Gu, Deng, and Su}]{Gu-Yu-ACL-2023-Pangu}
Yu~Gu, Xiang Deng, and Yu~Su. 2023.
\newblock \href {https://doi.org/10.18653/v1/2023.acl-long.270} {Don{'}t generate, discriminate: A proposal for grounding language models to real-world environments}.
\newblock In \emph{Proceedings of the 61st Annual Meeting of the Association for Computational Linguistics (Volume 1: Long Papers)}, pages 4928--4949, Toronto, Canada. Association for Computational Linguistics.

\bibitem[{Gu et~al.(2021)Gu, Kase, Vanni, Sadler, Liang, Yan, and Su}]{Gu-Yu-WWW-2021-GrailQA}
Yu~Gu, Sue Kase, Michelle Vanni, Brian Sadler, Percy Liang, Xifeng Yan, and Yu~Su. 2021.
\newblock \href {https://doi.org/10.1145/3442381.3449992} {Beyond i.i.d.: Three levels of generalization for question answering on knowledge bases}.
\newblock In \emph{Proceedings of the Web Conference 2021}, WWW '21, page 3477–3488, New York, NY, USA. Association for Computing Machinery.

\bibitem[{Gu et~al.(2024)Gu, Shu, Yu, Liu, Dong, Tang, Srinivasa, Latapie, and Su}]{Gu-Yu-arXiv-2024-Middleware}
Yu~Gu, Yiheng Shu, Hao Yu, Xiao Liu, Yuxiao Dong, Jie Tang, Jayanth Srinivasa, Hugo Latapie, and Yu~Su. 2024.
\newblock \href {http://arxiv.org/abs/2402.14672} {Middleware for llms: Tools are instrumental for language agents in complex environments}.

\bibitem[{Hu et~al.(2018)Hu, Zou, Yu, Wang, and Zhao}]{Hu-2018-TKDE-gAnswer}
Sen Hu, Lei Zou, Jeffrey~Xu Yu, Haixun Wang, and Dongyan Zhao. 2018.
\newblock \href {https://doi.org/10.1109/TKDE.2017.2766634} {Answering natural language questions by subgraph matching over knowledge graphs}.
\newblock \emph{IEEE Transactions on Knowledge and Data Engineering}, 30(5):824--837.

\bibitem[{Jiang et~al.(2023{\natexlab{a}})Jiang, Sablayrolles, Mensch, Bamford, Chaplot, de~las Casas, Bressand, Lengyel, Lample, Saulnier, Lavaud, Lachaux, Stock, Scao, Lavril, Wang, Lacroix, and Sayed}]{Jiang-Albert-Q-arXiv-2023-Mistral-7B}
Albert~Q. Jiang, Alexandre Sablayrolles, Arthur Mensch, Chris Bamford, Devendra~Singh Chaplot, Diego de~las Casas, Florian Bressand, Gianna Lengyel, Guillaume Lample, Lucile Saulnier, Lélio~Renard Lavaud, Marie-Anne Lachaux, Pierre Stock, Teven~Le Scao, Thibaut Lavril, Thomas Wang, Timothée Lacroix, and William~El Sayed. 2023{\natexlab{a}}.
\newblock \href {http://arxiv.org/abs/2310.06825} {Mistral 7b}.

\bibitem[{Jiang et~al.(2023{\natexlab{b}})Jiang, Zhou, Dong, Ye, Zhao, and Wen}]{Jiang-EMNLP-2023-StructGPT}
Jinhao Jiang, Kun Zhou, Zican Dong, Keming Ye, Xin Zhao, and Ji-Rong Wen. 2023{\natexlab{b}}.
\newblock \href {https://doi.org/10.18653/v1/2023.emnlp-main.574} {{S}truct{GPT}: A general framework for large language model to reason over structured data}.
\newblock In \emph{Proceedings of the 2023 Conference on Empirical Methods in Natural Language Processing}, pages 9237--9251, Singapore. Association for Computational Linguistics.

\bibitem[{Jiang et~al.(2024)Jiang, Zhou, Zhao, Song, Zhu, Zhu, and Wen}]{Jiang-Jinhao-arXiv-2024-KG-Agent}
Jinhao Jiang, Kun Zhou, Wayne~Xin Zhao, Yang Song, Chen Zhu, Hengshu Zhu, and Ji-Rong Wen. 2024.
\newblock \href {http://arxiv.org/abs/2402.11163} {Kg-agent: An efficient autonomous agent framework for complex reasoning over knowledge graph}.

\bibitem[{Jiang et~al.(2023{\natexlab{c}})Jiang, Zhou, Zhao, and Wen}]{Jinhao-Jiang-ICLR-2023-UniKGQA}
Jinhao Jiang, Kun Zhou, Xin Zhao, and Ji-Rong Wen. 2023{\natexlab{c}}.
\newblock \href {https://openreview.net/forum?id=Z63RvyAZ2Vh} {Uni{KGQA}: Unified retrieval and reasoning for solving multi-hop question answering over knowledge graph}.
\newblock In \emph{The Eleventh International Conference on Learning Representations}.

\bibitem[{Kwon et~al.(2023)Kwon, Li, Zhuang, Sheng, Zheng, Yu, Gonzalez, Zhang, and Stoica}]{Kwon-Woosuk-SOSP-2023-vLLM}
Woosuk Kwon, Zhuohan Li, Siyuan Zhuang, Ying Sheng, Lianmin Zheng, Cody~Hao Yu, Joseph Gonzalez, Hao Zhang, and Ion Stoica. 2023.
\newblock \href {https://doi.org/10.1145/3600006.3613165} {Efficient memory management for large language model serving with pagedattention}.
\newblock In \emph{Proceedings of the 29th Symposium on Operating Systems Principles}, SOSP '23, page 611–626, New York, NY, USA. Association for Computing Machinery.

\bibitem[{Lan et~al.(2023)Lan, He, Jiang, Jiang, Zhao, and Wen}]{Lan-Yunshi-TKDE-2023-ComplexKBQA-Survey}
Yunshi Lan, Gaole He, Jinhao Jiang, Jing Jiang, Wayne~Xin Zhao, and Ji-Rong Wen. 2023.
\newblock \href {https://doi.org/10.1109/TKDE.2022.3223858} {Complex knowledge base question answering: A survey}.
\newblock \emph{IEEE Transactions on Knowledge and Data Engineering}, 35(11):11196--11215.

\bibitem[{Lan and Jiang(2020)}]{Lan-Yunshi-ACL-2020-Query-Graph-Generation}
Yunshi Lan and Jing Jiang. 2020.
\newblock \href {https://doi.org/10.18653/v1/2020.acl-main.91} {Query graph generation for answering multi-hop complex questions from knowledge bases}.
\newblock In \emph{Proceedings of the 58th Annual Meeting of the Association for Computational Linguistics}, pages 969--974, Online. Association for Computational Linguistics.

\bibitem[{Li et~al.(2020)Li, Min, Iyer, Mehdad, and Yih}]{Li-Belinda-EMNLP-2020-ELQ}
Belinda~Z. Li, Sewon Min, Srinivasan Iyer, Yashar Mehdad, and Wen-tau Yih. 2020.
\newblock \href {https://doi.org/10.18653/v1/2020.emnlp-main.522} {Efficient one-pass end-to-end entity linking for questions}.
\newblock In \emph{Proceedings of the 2020 Conference on Empirical Methods in Natural Language Processing (EMNLP)}, pages 6433--6441, Online. Association for Computational Linguistics.

\bibitem[{Li et~al.(2023{\natexlab{a}})Li, Ma, Zhuang, Gu, Su, and Chen}]{Li-Tianle-ACL-2023-KB-BINDER}
Tianle Li, Xueguang Ma, Alex Zhuang, Yu~Gu, Yu~Su, and Wenhu Chen. 2023{\natexlab{a}}.
\newblock \href {https://doi.org/10.18653/v1/2023.acl-long.385} {Few-shot in-context learning on knowledge base question answering}.
\newblock In \emph{Proceedings of the 61st Annual Meeting of the Association for Computational Linguistics (Volume 1: Long Papers)}, pages 6966--6980, Toronto, Canada. Association for Computational Linguistics.

\bibitem[{Li et~al.(2023{\natexlab{b}})Li, Fan, Gu, Li, Duan, Dong, Liu, and Wang}]{Li-Zhenyu-arXiv-2023-FlexKBQA}
Zhenyu Li, Sunqi Fan, Yu~Gu, Xiuxing Li, Zhichao Duan, Bowen Dong, Ning Liu, and Jianyong Wang. 2023{\natexlab{b}}.
\newblock \href {http://arxiv.org/abs/2308.12060} {Flexkbqa: A flexible llm-powered framework for few-shot knowledge base question answering}.

\bibitem[{Lightman et~al.(2023)Lightman, Kosaraju, Burda, Edwards, Baker, Lee, Leike, Schulman, Sutskever, and Cobbe}]{Lightman-Hunter-arXiv-2023-ProcessSupervision}
Hunter Lightman, Vineet Kosaraju, Yura Burda, Harri Edwards, Bowen Baker, Teddy Lee, Jan Leike, John Schulman, Ilya Sutskever, and Karl Cobbe. 2023.
\newblock \href {http://arxiv.org/abs/2305.20050} {Let's verify step by step}.

\bibitem[{Liu et~al.(2024)Liu, Yu, Zhang, Xu, Lei, Lai, Gu, Ding, Men, Yang, Zhang, Deng, Zeng, Du, Zhang, Shen, Zhang, Su, Sun, Huang, Dong, and Tang}]{Liu-Xiao-ICLR-2023-AgentBench}
Xiao Liu, Hao Yu, Hanchen Zhang, Yifan Xu, Xuanyu Lei, Hanyu Lai, Yu~Gu, Hangliang Ding, Kaiwen Men, Kejuan Yang, Shudan Zhang, Xiang Deng, Aohan Zeng, Zhengxiao Du, Chenhui Zhang, Sheng Shen, Tianjun Zhang, Yu~Su, Huan Sun, Minlie Huang, Yuxiao Dong, and Jie Tang. 2024.
\newblock \href {https://openreview.net/forum?id=zAdUB0aCTQ} {Agentbench: Evaluating {LLM}s as agents}.
\newblock In \emph{The Twelfth International Conference on Learning Representations}.

\bibitem[{OpenAI(2023)}]{OpenAI-2023-GPT4}
OpenAI. 2023.
\newblock \href {https://arxiv.org/abs/2303.08774} {Gpt-4 technical report}.
\newblock \emph{ArXiv}, abs/2303.08774.

\bibitem[{Ouyang et~al.(2022)Ouyang, Wu, Jiang, Almeida, Wainwright, Mishkin, Zhang, Agarwal, Slama, Ray, Schulman, Hilton, Kelton, Miller, Simens, Askell, Welinder, Christiano, Leike, and Lowe}]{Ouyang-Long-NeurIPS-2022-InstructGPT}
Long Ouyang, Jeffrey Wu, Xu~Jiang, Diogo Almeida, Carroll Wainwright, Pamela Mishkin, Chong Zhang, Sandhini Agarwal, Katarina Slama, Alex Ray, John Schulman, Jacob Hilton, Fraser Kelton, Luke Miller, Maddie Simens, Amanda Askell, Peter Welinder, Paul~F Christiano, Jan Leike, and Ryan Lowe. 2022.
\newblock \href {https://proceedings.neurips.cc/paper_files/paper/2022/file/b1efde53be364a73914f58805a001731-Paper-Conference.pdf} {Training language models to follow instructions with human feedback}.
\newblock In \emph{Advances in Neural Information Processing Systems}, volume~35, pages 27730--27744. Curran Associates, Inc.

\bibitem[{Raffel et~al.(2020)Raffel, Shazeer, Roberts, Lee, Narang, Matena, Zhou, Li, and Liu}]{Raffel-JMLR-2020-T5}
Colin Raffel, Noam Shazeer, Adam Roberts, Katherine Lee, Sharan Narang, Michael Matena, Yanqi Zhou, Wei Li, and Peter~J. Liu. 2020.
\newblock Exploring the limits of transfer learning with a unified text-to-text transformer.
\newblock \emph{J. Mach. Learn. Res.}, 21(1).

\bibitem[{Rasley et~al.(2020)Rasley, Rajbhandari, Ruwase, and He}]{Rasley-Jeff-KDD-2020-DeepSpeed}
Jeff Rasley, Samyam Rajbhandari, Olatunji Ruwase, and Yuxiong He. 2020.
\newblock \href {https://doi.org/10.1145/3394486.3406703} {Deepspeed: System optimizations enable training deep learning models with over 100 billion parameters}.
\newblock In \emph{Proceedings of the 26th ACM SIGKDD International Conference on Knowledge Discovery \& Data Mining}, KDD '20, page 3505–3506, New York, NY, USA. Association for Computing Machinery.

\bibitem[{Shu et~al.(2022)Shu, Yu, Li, Karlsson, Ma, Qu, and Lin}]{Shu-Yiheng-EMNLP-2022-TIARA}
Yiheng Shu, Zhiwei Yu, Yuhan Li, B{\"o}rje Karlsson, Tingting Ma, Yuzhong Qu, and Chin-Yew Lin. 2022.
\newblock \href {https://doi.org/10.18653/v1/2022.emnlp-main.555} {{TIARA}: Multi-grained retrieval for robust question answering over large knowledge base}.
\newblock In \emph{Proceedings of the 2022 Conference on Empirical Methods in Natural Language Processing}, pages 8108--8121, Abu Dhabi, United Arab Emirates. Association for Computational Linguistics.

\bibitem[{Su et~al.(2023)Su, Lu, Pan, Murtadha, Wen, and Liu}]{Su-Jianlin-arXiv-2023-RoPE}
Jianlin Su, Yu~Lu, Shengfeng Pan, Ahmed Murtadha, Bo~Wen, and Yunfeng Liu. 2023.
\newblock \href {http://arxiv.org/abs/2104.09864} {Roformer: Enhanced transformer with rotary position embedding}.

\bibitem[{Sun et~al.(2024)Sun, Xu, Tang, Wang, Lin, Gong, Shum, and Guo}]{Jiashuo-Sun-ICLR-2024-Think-on-Graph}
Jiashuo Sun, Chengjin Xu, Lumingyuan Tang, Saizhuo Wang, Chen Lin, Yeyun Gong, Heung-Yeung Shum, and Jian Guo. 2024.
\newblock \href {https://openreview.net/forum?id=nnVO1PvbTv} {Think-on-graph: Deep and responsible reasoning of large language model on knowledge graph}.
\newblock In \emph{The Twelfth International Conference on Learning Representations}.

\bibitem[{Talmor and Berant(2018)}]{Talmor-Alon-NAACL-2018-ComplexWebQuestions-CWQ}
Alon Talmor and Jonathan Berant. 2018.
\newblock \href {https://doi.org/10.18653/v1/N18-1059} {The web as a knowledge-base for answering complex questions}.
\newblock In \emph{Proceedings of the 2018 Conference of the North {A}merican Chapter of the Association for Computational Linguistics: Human Language Technologies, Volume 1 (Long Papers)}, pages 641--651, New Orleans, Louisiana. Association for Computational Linguistics.

\bibitem[{Touvron et~al.(2023)Touvron, Martin, Stone, Albert, Almahairi, Babaei, Bashlykov, Batra, Bhargava, Bhosale, Bikel, Blecher, Ferrer, Chen, Cucurull, Esiobu, Fernandes, Fu, Fu, Fuller, Gao, Goswami, Goyal, Hartshorn, Hosseini, Hou, Inan, Kardas, Kerkez, Khabsa, Kloumann, Korenev, Koura, Lachaux, Lavril, Lee, Liskovich, Lu, Mao, Martinet, Mihaylov, Mishra, Molybog, Nie, Poulton, Reizenstein, Rungta, Saladi, Schelten, Silva, Smith, Subramanian, Tan, Tang, Taylor, Williams, Kuan, Xu, Yan, Zarov, Zhang, Fan, Kambadur, Narang, Rodriguez, Stojnic, Edunov, and Scialom}]{Touvron-Hugo-arXiv-2023-Llama2}
Hugo Touvron, Louis Martin, Kevin Stone, Peter Albert, Amjad Almahairi, Yasmine Babaei, Nikolay Bashlykov, Soumya Batra, Prajjwal Bhargava, Shruti Bhosale, Dan Bikel, Lukas Blecher, Cristian~Canton Ferrer, Moya Chen, Guillem Cucurull, David Esiobu, Jude Fernandes, Jeremy Fu, Wenyin Fu, Brian Fuller, Cynthia Gao, Vedanuj Goswami, Naman Goyal, Anthony Hartshorn, Saghar Hosseini, Rui Hou, Hakan Inan, Marcin Kardas, Viktor Kerkez, Madian Khabsa, Isabel Kloumann, Artem Korenev, Punit~Singh Koura, Marie-Anne Lachaux, Thibaut Lavril, Jenya Lee, Diana Liskovich, Yinghai Lu, Yuning Mao, Xavier Martinet, Todor Mihaylov, Pushkar Mishra, Igor Molybog, Yixin Nie, Andrew Poulton, Jeremy Reizenstein, Rashi Rungta, Kalyan Saladi, Alan Schelten, Ruan Silva, Eric~Michael Smith, Ranjan Subramanian, Xiaoqing~Ellen Tan, Binh Tang, Ross Taylor, Adina Williams, Jian~Xiang Kuan, Puxin Xu, Zheng Yan, Iliyan Zarov, Yuchen Zhang, Angela Fan, Melanie Kambadur, Sharan Narang, Aurelien Rodriguez, Robert Stojnic, Sergey Edunov, and Thomas Scialom. 2023.
\newblock \href {http://arxiv.org/abs/2307.09288} {Llama 2: Open foundation and fine-tuned chat models}.

\bibitem[{Vrande\v{c}i\'{c} and Kr\"{o}tzsch(2014)}]{Vrandečić-Denny-CACM-2014-Wikidata}
Denny Vrande\v{c}i\'{c} and Markus Kr\"{o}tzsch. 2014.
\newblock \href {https://doi.org/10.1145/2629489} {Wikidata: a free collaborative knowledgebase}.
\newblock \emph{Commun. ACM}, 57(10):78–85.

\bibitem[{Wang et~al.(2023)Wang, Wei, Schuurmans, Le, Chi, Narang, Chowdhery, and Zhou}]{Wang-Xuezhi-ICLR-2023-SelfConsistency}
Xuezhi Wang, Jason Wei, Dale Schuurmans, Quoc~V Le, Ed~H. Chi, Sharan Narang, Aakanksha Chowdhery, and Denny Zhou. 2023.
\newblock \href {https://openreview.net/forum?id=1PL1NIMMrw} {Self-consistency improves chain of thought reasoning in language models}.
\newblock In \emph{The Eleventh International Conference on Learning Representations}.

\bibitem[{Wei et~al.(2022)Wei, Wang, Schuurmans, Bosma, brian ichter, Xia, Chi, Le, and Zhou}]{Wei-Jason-NeurIPS-2022-ChainOfThought}
Jason Wei, Xuezhi Wang, Dale Schuurmans, Maarten Bosma, brian ichter, Fei Xia, Ed~H. Chi, Quoc~V Le, and Denny Zhou. 2022.
\newblock \href {https://openreview.net/forum?id=_VjQlMeSB_J} {Chain of thought prompting elicits reasoning in large language models}.
\newblock In \emph{Advances in Neural Information Processing Systems}.

\bibitem[{Yan et~al.(2021)Yan, Li, Wang, Zhang, Daoguang, Zhang, Wu, and Xu}]{Yan-Yuanmeng-EMNLP-2021-Large-scale-relation-learning}
Yuanmeng Yan, Rumei Li, Sirui Wang, Hongzhi Zhang, Zan Daoguang, Fuzheng Zhang, Wei Wu, and Weiran Xu. 2021.
\newblock \href {https://doi.org/10.18653/v1/2021.emnlp-main.296} {Large-scale relation learning for question answering over knowledge bases with pre-trained language models}.
\newblock In \emph{Proceedings of the 2021 Conference on Empirical Methods in Natural Language Processing}, pages 3653--3660, Online and Punta Cana, Dominican Republic. Association for Computational Linguistics.

\bibitem[{Ye et~al.(2022)Ye, Yavuz, Hashimoto, Zhou, and Xiong}]{Ye-Xi-ACL-2022-RNG-KBQA}
Xi~Ye, Semih Yavuz, Kazuma Hashimoto, Yingbo Zhou, and Caiming Xiong. 2022.
\newblock \href {https://doi.org/10.18653/v1/2022.acl-long.417} {{RNG}-{KBQA}: Generation augmented iterative ranking for knowledge base question answering}.
\newblock In \emph{Proceedings of the 60th Annual Meeting of the Association for Computational Linguistics (Volume 1: Long Papers)}, pages 6032--6043, Dublin, Ireland. Association for Computational Linguistics.

\bibitem[{Yih et~al.(2015)Yih, Chang, He, and Gao}]{Yih-Wen-tau-ACL-2015-STAGG}
Wen-tau Yih, Ming-Wei Chang, Xiaodong He, and Jianfeng Gao. 2015.
\newblock \href {https://doi.org/10.3115/v1/P15-1128} {Semantic parsing via staged query graph generation: Question answering with knowledge base}.
\newblock In \emph{Proceedings of the 53rd Annual Meeting of the Association for Computational Linguistics and the 7th International Joint Conference on Natural Language Processing (Volume 1: Long Papers)}, pages 1321--1331, Beijing, China. Association for Computational Linguistics.

\bibitem[{Yih et~al.(2016)Yih, Richardson, Meek, Chang, and Suh}]{Yih-Wen-tau-ACL-2016-WebQSP}
Wen-tau Yih, Matthew Richardson, Chris Meek, Ming-Wei Chang, and Jina Suh. 2016.
\newblock \href {https://doi.org/10.18653/v1/P16-2033} {The value of semantic parse labeling for knowledge base question answering}.
\newblock In \emph{Proceedings of the 54th Annual Meeting of the Association for Computational Linguistics (Volume 2: Short Papers)}, pages 201--206, Berlin, Germany. Association for Computational Linguistics.

\bibitem[{Yu et~al.(2023)Yu, Zhang, Ng, Zhu, Li, Wang, Hu, Wang, Wang, and Xiang}]{Yu-Donghan-ICLR-2023-DecAF}
Donghan Yu, Sheng Zhang, Patrick Ng, Henghui Zhu, Alexander~Hanbo Li, Jun Wang, Yiqun Hu, William~Yang Wang, Zhiguo Wang, and Bing Xiang. 2023.
\newblock \href {https://openreview.net/forum?id=XHc5zRPxqV9} {Dec{AF}: Joint decoding of answers and logical forms for question answering over knowledge bases}.
\newblock In \emph{The Eleventh International Conference on Learning Representations}.

\bibitem[{Zaheer et~al.(2020)Zaheer, Guruganesh, Dubey, Ainslie, Alberti, Ontanon, Pham, Ravula, Wang, Yang, and Ahmed}]{Zaheer-Manzil-NIPS-2020-BigBird}
Manzil Zaheer, Guru Guruganesh, Avinava Dubey, Joshua Ainslie, Chris Alberti, Santiago Ontanon, Philip Pham, Anirudh Ravula, Qifan Wang, Li~Yang, and Amr Ahmed. 2020.
\newblock Big bird: transformers for longer sequences.
\newblock In \emph{Proceedings of the 34th International Conference on Neural Information Processing Systems}, NIPS'20, Red Hook, NY, USA. Curran Associates Inc.

\bibitem[{Zhang et~al.(2022{\natexlab{a}})Zhang, Zhang, Yu, Tang, Tang, Li, and Chen}]{Zhang-Jing-ACL-2022-Subgraph-Retrieval-Enhanced-Model}
Jing Zhang, Xiaokang Zhang, Jifan Yu, Jian Tang, Jie Tang, Cuiping Li, and Hong Chen. 2022{\natexlab{a}}.
\newblock \href {https://doi.org/10.18653/v1/2022.acl-long.396} {Subgraph retrieval enhanced model for multi-hop knowledge base question answering}.
\newblock In \emph{Proceedings of the 60th Annual Meeting of the Association for Computational Linguistics (Volume 1: Long Papers)}, pages 5773--5784, Dublin, Ireland. Association for Computational Linguistics.

\bibitem[{Zhang et~al.(2023)Zhang, Zhang, Wang, Cao, Huang, Li, Chen, and Li}]{Zhang-Lingxi-ACL-2023-FC-KBQA}
Lingxi Zhang, Jing Zhang, Yanling Wang, Shulin Cao, Xinmei Huang, Cuiping Li, Hong Chen, and Juanzi Li. 2023.
\newblock \href {https://doi.org/10.18653/v1/2023.acl-long.57} {{FC}-{KBQA}: A fine-to-coarse composition framework for knowledge base question answering}.
\newblock In \emph{Proceedings of the 61st Annual Meeting of the Association for Computational Linguistics (Volume 1: Long Papers)}, pages 1002--1017, Toronto, Canada. Association for Computational Linguistics.

\bibitem[{Zhang et~al.(2022{\natexlab{b}})Zhang, Jin, Li, and Wang}]{Zhang-Yanan-Edge-aware-GNN-Neurosci-2022}
Yanan Zhang, Li~Jin, Xiaoyu Li, and Honqi Wang. 2022{\natexlab{b}}.
\newblock Edge-aware graph neural network for multi-hop path reasoning over knowledge base.
\newblock \emph{Comput. Intell. Neurosci.}, 2022:4734179.

\bibitem[{Zhang et~al.(2018)Zhang, Dai, Kozareva, Smola, and Song}]{Zhang-Yuyu-AAAI-2018-MetaQA}
Yuyu Zhang, Hanjun Dai, Zornitsa Kozareva, Alexander~J. Smola, and Le~Song. 2018.
\newblock Variational reasoning for question answering with knowledge graph.
\newblock In \emph{Proceedings of the Thirty-Second AAAI Conference on Artificial Intelligence and Thirtieth Innovative Applications of Artificial Intelligence Conference and Eighth AAAI Symposium on Educational Advances in Artificial Intelligence}, AAAI'18/IAAI'18/EAAI'18. AAAI Press.

\bibitem[{Zhu et~al.(2023)Zhu, Yang, Wang, Song, Wu, Wei, and Li}]{Zhu-Dawei-arXiv-2023-PoSE}
Dawei Zhu, Nan Yang, Liang Wang, Yifan Song, Wenhao Wu, Furu Wei, and Sujian Li. 2023.
\newblock \href {http://arxiv.org/abs/2309.10400} {Pose: Efficient context window extension of llms via positional skip-wise training}.

\bibitem[{Zong et~al.(2024)Zong, Yan, Lu, Shao, Huang, Chang, and Zhuang}]{Zong-Chang-arXiv-2024-Triad}
Chang Zong, Yuchen Yan, Weiming Lu, Jian Shao, Eliot Huang, Heng Chang, and Yueting Zhuang. 2024.
\newblock \href {http://arxiv.org/abs/2402.14320} {Triad: A framework leveraging a multi-role llm-based agent to solve knowledge base question answering}.

\end{thebibliography}
\appendix
\section{Appendix}

This appendix provides detailed experimental results and offers further discussion.

\subsection{Additional Statistics of Datasets \& Databases}

\begin{table}[ht]
\centering
\scalebox{1}{
    \begin{tabular}{ccc}
    \hline
    \textbf{Database} & \textbf{\#Node}     & \textbf{\#Rel}  \\ \hline
    Freebase & 22,767,149 & 20,815 \\
    Wikidata & 17,754     & 1,267  \\
    Movie KG & 38,131     & 9      \\ \hline
    \end{tabular}
}
\caption{Statistics of the databases.}
\label{app_tab:db_statistics}
\end{table}

\begin{table}[ht]
\scalebox{1}{
    \begin{tabular}{cc}
    \hline
    \textbf{Question Type} & \textbf{Dist. (Train/Dev/Test)} \\ \hline
    conjunction & 42.00 / 43.59 / 44.60 \\
    composition & 47.27 / 44.76 / 43.78 \\
    comparative & 5.54 / 6.22 / 6.03 \\
    superlative & 5.19 / 5.43 / 5.58 \\ \hline
    \end{tabular}
}
\caption{Distribution (Dist.) of question types in the original CWQ.}
\label{app_tab:distribution_qtype_cwq}
\end{table}

Table \ref{app_tab:db_statistics} presents the statistical information for Freebase, Wikidata, and Movie KG. For Freebase, we utilize a script\footnote{\url{https://github.com/lanyunshi/Multi-hopComplexKBQA/blob/master/code/FreebaseTool/FilterEnglishTriplets.py}} to eliminate non-english triples. In the case of Wikidata, we refer to the subset processed by KQA Pro \cite{Cao-Shulin-ACL-2022-KQAPro}. Meanwhile, Movie KG is derived by converting the knowledge base provided by MetaQA into RDF triples.
Additionally, Table \ref{app_tab:distribution_qtype_cwq} outlines the distribution of question types in the original CWQ.

\subsection{System Configurations}

\subsubsection{Implementation Details}
\label{app_sec:imp_details}

In the interactive process, our study invokes the OpenAI GPT4-Turbo (\texttt{gpt-4-1106-preview}) API to serve as a LLM agent. For each type of questions, we selected and annotated two cases with complete interactive processes as exemplars for in-context learning. During the inference stage, for the WebQSP and MetaQA datasets, all annotated demonstrations are employed as few-shot learning examples, specifically, 4-shot for WebQSP and 6-shot for MetaQA. For CWQ and KQA Pro, we train a \texttt{bert-base-uncased} \footnote{\url{https://huggingface.co/bert-base-uncased}} model based on low-resource training data as a question type classifier (cls), and then select two exemplars based on the predicted question type, namely cls+2-shot for both CWQ and KQA Pro.

In the development of the \textbf{SearchNodes} tool, Elasticsearch\footnote{\url{https://github.com/elastic/elasticsearch}} is employed to extract all node surface names from the Freebase and MetaQA databases, and vector search techniques are implemented to perform queries on nodes within Wikidata. 
For the ranking algorithm of the \textbf{SearchGraphPatterns} tool, vector retrieval methods are similarly employed. All processes related to vectorization utilize the OpenAI \texttt{text-embedding-ada-002} API to generate vectors and employ Chroma\footnote{\url{https://github.com/chroma-core/chroma}} for indexing and searching.
Moreover, for the functionality of the \textbf{ExecuteSPARQL} tool, Virtuoso\footnote{\url{https://github.com/openlink/virtuoso-opensource}} serves as the underlying graph query engine.

In the process of fine-tuning open-source LLMs, we utilize, \texttt{Mistral-7B-v0.1} \cite{Jiang-Albert-Q-arXiv-2023-Mistral-7B} and \texttt{Llama-2-13b} \cite{Touvron-Hugo-arXiv-2023-Llama2}. For training optimization, DeepSpeed \cite{Rasley-Jeff-KDD-2020-DeepSpeed} is employed, while inference tasks are accelerated using vLLM \cite{Kwon-Woosuk-SOSP-2023-vLLM}. 
\texttt{Llama-2-13b} is trained on four NVIDIA A100 80GB GPUs, the other experiments are conducted on two NVIDIA A100 80GB GPUs.

\subsubsection{Hyper-Parameter Setting}
\label{app_sec:hyper_setting}

\begin{table}[ht]
\centering
    \begin{tabular}{lc}
    \hline
    \textbf{Parameter}   & \textbf{Value} \\ \hline
    temperature & 0.7 \\
    top\_p      & 1 \\
    n           & 6 \# 1 when annotating data \\
    stop & {[}"\textbackslash{}nObservation", "\textbackslash{}nThought"{]} \\
    max\_tokens & 384 \\ \hline
    \end{tabular}
\caption{Assignments of hyper-parameters for inference.}
\label{app_tab:hyper_infer}
\end{table}

\begin{table}[ht]
\centering
    \begin{tabular}{lc}
    \hline
    \textbf{Parameter} & \textbf{Value} \\ \hline
    batch size (per GPU)        & 1     \\
    model max length            & 4096  \\
    learning rate               & 1e-5  \\
    weight decay                & 0.001 \\
    epochs                      & 10    \\
    warm up steps               & 0     \\
    gradient accumulation steps & 4     \\
    zero stage                  & 3     \\ \hline
    \end{tabular}
\caption{Assignments of hyper-parameters for fine-tuning open-source LLMs.}
\label{app_tab:hyper_deepspeed}
\end{table}

Table \ref{app_tab:hyper_infer} presents the parameter configurations for invoking the OpenAI API and fine-tuning open-source LLMs. 
We employ DeepSpeed \cite{Rasley-Jeff-KDD-2020-DeepSpeed} to efficiently fine-tune open-source LLMs. The hyperparameter settings utilized for training are detailed in Table \ref{app_tab:hyper_deepspeed}.

For both the SearchNodes and SearchGraphPatterns tools, the number of returned results is set to 10. 
In the interaction process, the maximum number of turns is limited to 20, corresponding to 10 rounds.

\subsubsection{Instruction Text}
\label{app_sec:instruction_text}

\begin{figure*}[hp!]
  \centering
  \begin{tikzpicture}
    \node[inner sep=2pt] (image) at (0,0) {
      \includegraphics[width=1\textwidth,page=1]{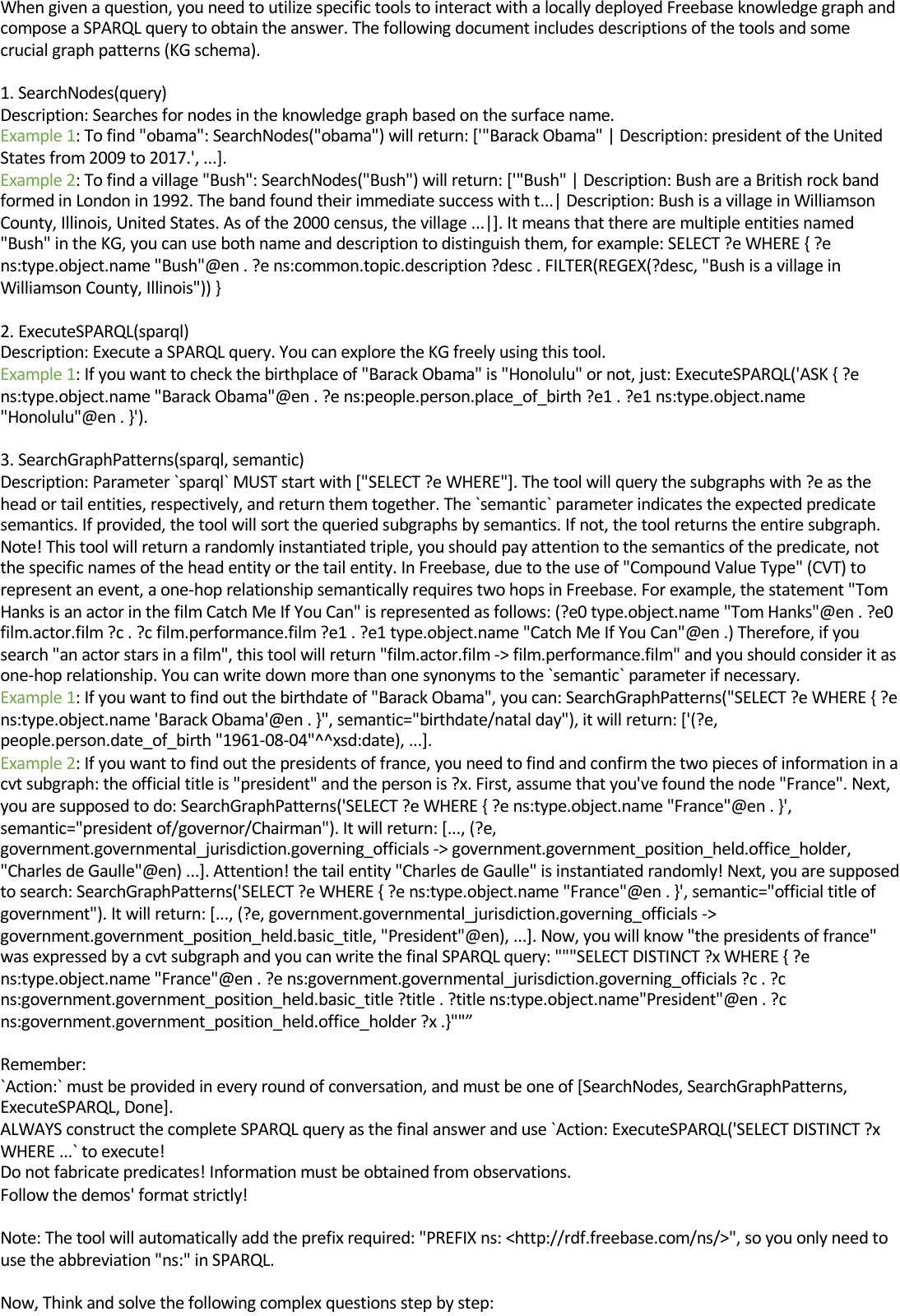}
    };
    \draw[black, thin] (image.south west) rectangle (image.north east);
  \end{tikzpicture}
  \caption{Instruction text of Freebase.}
  \label{app_fig:inst_fb}
\end{figure*}

\begin{figure*}[hp!]
  \centering
  \begin{tikzpicture}
    \node[inner sep=2pt] (image) at (0,0) {
      \includegraphics[width=1\textwidth,page=2]{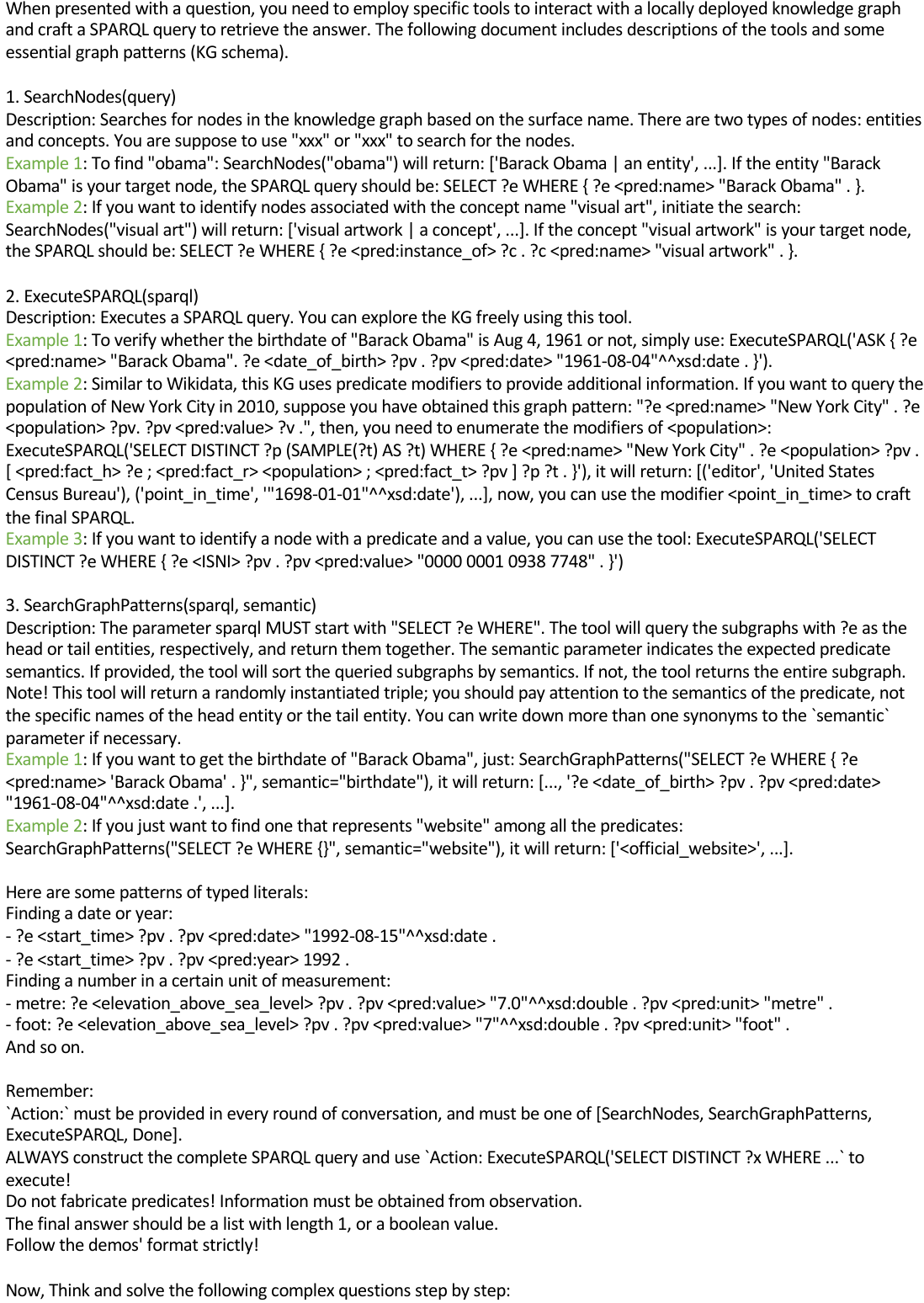}
    };
    \draw[black, thin] (image.south west) rectangle (image.north east);
  \end{tikzpicture}
  \caption{Instruction text of Wikidata.}
  \label{app_fig:inst_wikidata}
\end{figure*}

\begin{figure*}[hp!]
  \centering
  \begin{tikzpicture}
    \node[inner sep=2pt] (image) at (0,0) {
      \includegraphics[width=1\textwidth,page=3]{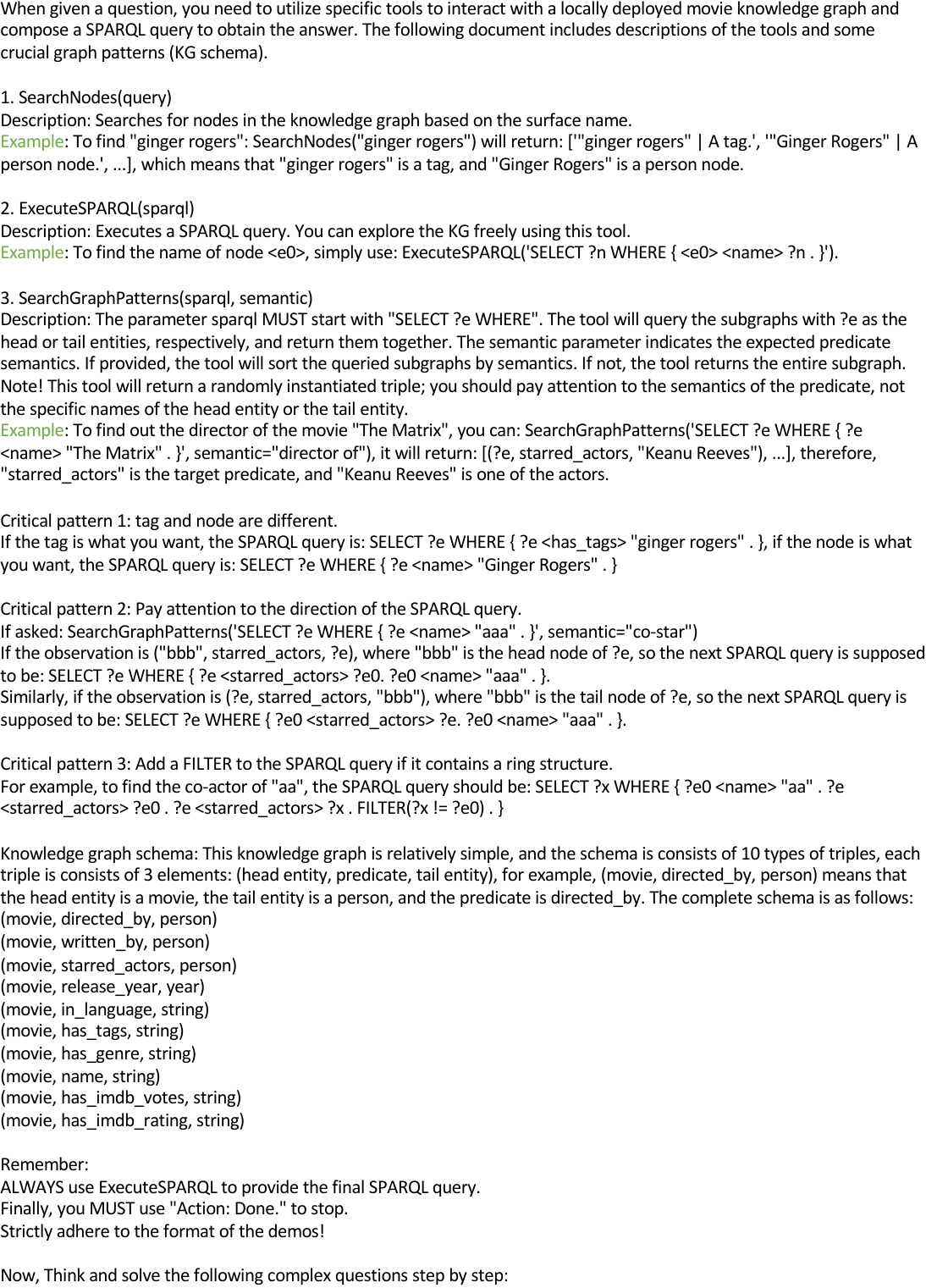}
    };
    \draw[black, thin] (image.south west) rectangle (image.north east);
  \end{tikzpicture}
  \caption{Instruction text of Movie KG for MetaQA.}
  \label{app_fig:inst_metaqa}
\end{figure*}

Instruction text consists of tool descriptions, tool usages and the format of interaction.
The tools designed in this work is capable of processing a variety of complex questions through a unified approach on the different databases.
Figures \ref{app_fig:inst_fb}, \ref{app_fig:inst_wikidata}, and \ref{app_fig:inst_metaqa} showcase comprehensive instruction texts for Freebase, Wikidata, and Movie KG, respectively. 
For further insight into the tool's applications, examples within these figures are highlighted in green.

\subsection{Additional Results of Entity Linking}

\begin{table*}[htp]
\centering
    \begin{tabular}{lcccc}
    \hline
    \multicolumn{5}{c}{\textbf{WebQSP}}                                                   \\ \hline
    \textbf{Methods}       & \textbf{1-hop} & \textbf{2-hop} & \textbf{Overall} & \textbf{RHits@1} \\ \hline
    Ours          & 69.99          & 72.41          & 71.20            & 72.47            \\
    Ours w/golden & 77.50          & 79.79          & 78.64            & 79.25            \\
    Gain          & 7.51           & 7.38           & 7.44             & 6.77             \\ \hline
    \end{tabular}
\caption{The impact of given golden entities on WebQSP by question type}
\label{app_tab:el_res_full_webqsp}
\end{table*}

\begin{table*}[ht]
\centering
    \begin{tabular}{lcccccc}
    \hline
    \multicolumn{7}{c}{\textbf{CWQ}}                                                                        \\ \hline
    \textbf{Methods}              & \textbf{Conj} & \textbf{Compo} & \textbf{Compa} & \textbf{Super} & \textbf{Overall} & \textbf{EM}    \\ \hline
    Ours                 & 47.44       & 59.00       & 47.89       & 41.96       & 49.07   & 59.17 \\
    Ours w/golden entity & 53.36       & 68.19       & 52.73       & 52.69       & 56.74   & 66.50 \\
    Gain                 & 5.92        & 9.19        & 4.84        & 10.72       & 7.67    & 7.33  \\ \hline
    \end{tabular}
\caption{The impact of given golden entities on CWQ by question type}
\label{app_tab:el_res_full_cwq}
\end{table*}

In Table \ref{app_tab:el_res_full_webqsp} and Table \ref{app_tab:el_res_full_cwq}, we report the impact of given golden entities on model performance by question type for WebQSP and CWQ, respectively.

\subsection{Additional Results of Exemplars Selection}

\begin{table*}[ht]
\centering
    \begin{tabular}{ccccccl}
    \hline
    \multicolumn{7}{c}{\textbf{CWQ}}                                                         \\ \hline
    \textbf{Setting} & \textbf{Conj} & \textbf{Compo} & \textbf{Compa} & \textbf{Super} & \textbf{Overall} & \multicolumn{1}{c}{\textbf{Ave Price}} \\ \hline
    \multicolumn{1}{l}{Ours (cls+2-shot)} & 58.64 & 66.71 & 41.35 & 50.79 & 54.69 & \$0.50 \\
    \multicolumn{1}{r}{w/4-shot}            & 38.02 & 51.32 & 75.66 & 63.37 & 57.19 & \$0.70 \\
    \multicolumn{1}{r}{zero-shot}           & 75.86 & 52.92 & 32.86 & 44.00 & 51.83 & \$0.37 \\ \hline
    \end{tabular}
\caption{The impact of exemplar selection on CWQ by question type}
\label{app_tab:shot_impact_cwq}
\end{table*}

\begin{table*}[ht]
\centering
    \begin{tabular}{rcccccc}
    \hline
    \multicolumn{7}{c}{\textbf{KQA Pro}} \\ \hline
    \multicolumn{1}{c}{\textbf{Setting}} & \textbf{Ct}  & \textbf{QA} & \textbf{QAQ} & \textbf{QN} & \textbf{QR}      & \\ \hline
    \multicolumn{1}{l}{Ours (cls+2-shot)} & 50 & 90 & 60 & 60 & 80    &        \\
    w/4-shot $\alpha$ & 20 & 70 & 40 & 40 & 50    &        \\
    w/4-shot $\beta$ & 50 & 70 & 50 & 70 & 20    &        \\
    zero-shot & 18 & 64 & 9  & 55 & 0     &        \\ \hline
    \multicolumn{1}{l}{(continued)}      & \textbf{QRQ} & \textbf{SA} & \textbf{SB}  & \textbf{VF} & \textbf{Overall} & \textbf{Ave Price} \\ \hline
    \multicolumn{1}{l}{Ours (cls+2-shot)} & 50 & 70 & 80 & 70 & 67.78 & \$0.38 \\
    w/4-shot $\alpha$ & 50 & 50 & 10 & 90 & 46.67 & \$0.50 \\
    w/4-shot $\beta$ & 20 & 80 & 40 & 70 & 52.22 & \$0.55 \\
    zero-shot & 0  & 45 & 9  & 27 & 25    & \$0.30 \\ \hline
    \end{tabular}
\caption{The impact of exemplar selection on KQA Pro by question type}
\label{app_tab:shot_impact_kqa}
\end{table*}

In Table \ref{app_tab:shot_impact_cwq} and Table \ref{app_tab:shot_impact_kqa}, we report the impact of exemplar selection on model performance by question type for CWQ and KQA Pro, respectively. This examination delineates how the choice of exemplars influences the effectiveness of the model across question types, underscoring the significance of tailored exemplar selection in prompt engineering.

% add (Reviewer F8iA 3)
The observation shows that the performance of our proposed approach on KQA Pro drops significantly in zero-shot compared to few-shot scenarios as observed in Table \ref{app_tab:shot_impact_kqa}. This discrepancy arises primarily due to the intricate nature of the questions within the KQA Pro dataset, which involve multi-hop reasoning, constraints, and qualifiers. Constructed using SPARQL templates and phrased naturally by human annotators, these questions are complex. In zero-shot settings, without exemplars, even human annotators may struggle with semantic parsing due to the varied and unseen KG schema. Conversely, the few-shot approach leverages a small set of carefully annotated examples, effectively guiding the parsing process and accounting for the observed performance differences.

\subsection{Interaction Turns and Costs}
\label{app_sec:round_price}

\begin{table*}
\centering
\scalebox{1}{
    \begin{tabular}{lcccc}
    \hline
    \multicolumn{1}{c}{\multirow{2}{*}{\textbf{Method}}} & \multicolumn{2}{c}{\textbf{WebQSP}} & \multicolumn{2}{c}{\textbf{CWQ}} \\ \cline{2-5} 
    \multicolumn{1}{c}{} & \textbf{Overall} & \textbf{Ave Price} & \textbf{Overall} & \textbf{Ave Price} \\ \hline
    Human Anno & 0.97 / 10.26 & \$0.16 & \multicolumn{2}{c}{1.34 / 11.59} \\ \hline
    Ours w/GPT-4 & 94.33 / 11.01 & \$0.39 & 80.50 / 14.57 & \$0.49 \\
    Ours w/GPT-4 + Golden & 95.33 / 9.17 & \$0.24 & 82.50 / 12.74 & \$0.38 \\
    Ours w/FT-Mistral & 74.05 / 11.71 & - & 64.67 / 13.29 & - \\ \hline
    \multicolumn{1}{c}{\multirow{2}{*}{\textbf{Method}}} & \multicolumn{2}{c}{\textbf{KQA Pro}} & \multicolumn{2}{c}{\textbf{MetaQA}} \\ \cline{2-5} 
    \multicolumn{1}{c}{} & \textbf{Overall} & \textbf{Ave Price} & \textbf{Overall} & \textbf{Ave Price} \\ \hline
    Human Anno & 0.80 / 11.75 & \$0.23 & 0.08 / 7.13 & \$0.06 \\ \hline
    Ours w/GPT-4 & 93.89 / 11.91 & \$0.33 & 99.67 / 7.45 & \$0.14 \\
    Ours w/FT-Mistral & 81.20 / 11.18 & - & 98.78 / 7.07 & - \\ \hline
    \end{tabular}
}
\caption{Comparison of turn count and cost across human-annotated data (Human Anno) and four datasets. For Human Anno, the average number of human interventions and interaction turn are reported. Concerning model performance, the average success rate and interaction turn are reported.}
\label{app_tab:info_dialog_turn_overall}
\end{table*}

In the process of interaction, the turn constitutes the fundamental unit, with each utterance made by any participant being recorded as a distinct turn.
Table \ref{app_tab:info_dialog_turn_overall} provides a statistical analysis of the interaction turns. A case is deemed successful if the LLM agent explicitly generates \texttt{Action:Done}. Otherwise, the dialogue is concluded once the maximum turn number is reached. 
It is feasible to consider the turn count of human-annotated data as a lower bound. 
It is observed that, except for on CWQ, GPT-4 have approached this lower bound. 
Specifically, in CWQ, the overall success rate per turn for GPT-4 is 80.5/14.57, which, compared to WebQSP, shows a decrease of 14.66\% and an increase of 39.96\% in terms of success rate and turn count, respectively. This also delineates the difficulty level of the dataset from one perspective.
In contrast, the KQA Pro dataset exhibits a stronger regularity, hence presenting a relatively lower level of difficulty. 
Furthermore, we discovered that on complicated question datasets, the average number of turns for fine-tuned LLMs is less than that for GPT-4, indicating that our annotated data effectively reduced the model's trial-and-error process by providing valuable information.
Additionally, we report the average cost per dialogue for reference \footnote{The price of gpt-4-1106-preview: Input \$0.01 / 1K tokens and Output \$0.03 / 1K tokens}. Given that we set the return sequence parameter to 6 for inference and to 1 for human annotation, the cost comparison is not strictly equivalent.
Complete results categorized by question type are available in the Table \ref{app_tab:info_dialog_turn_full_cwq}, Table \ref{app_tab:info_dialog_turn_full_kqa}, and Table \ref{app_tab:info_dialog_turn_full_metaqa}.

\begin{table*}[ht]
\centering
    \begin{tabular}{lcccc}
    \hline
    \multicolumn{1}{c}{\multirow{2}{*}{\textbf{Method}}} & \multicolumn{4}{c}{\textbf{WebQSP}}                                     \\ \cline{2-5} 
    \multicolumn{1}{c}{}                                 & \textbf{1-hop} & \textbf{2-hop} & \textbf{Overall} & \textbf{Ave Price} \\ \hline
    Human Anno            & 1.08 / 10.04  & 0.86 / 10.48  & 0.97 / 10.26  & \$0.16 \\ \hline
    Ours w/GPT-4          & 94.67 / 10.83 & 94.00 / 11.19 & 94.33 / 11.01 & \$0.39 \\
    Ours w/GPT-4 + Golden & 98.00 / 8.47  & 92.67 / 9.87  & 95.33 / 9.17  & \$0.24 \\
    Ours w/FT Mistral-7B  & 77.86 / 11.36 & 70.47 / 12.04 & 74.05 / 11.71 & -      \\ \hline
    \end{tabular}
\caption{Comparison of turn count and cost across on WebQSP by question type.}
\label{app_tab:info_dialog_turn_full_webqsp}
\end{table*}

\begin{table*}[ht]
\centering
\scalebox{0.85}{
    \begin{tabular}{lcccccc}
    \hline
    \multicolumn{1}{c}{\multirow{2}{*}{\textbf{Method}}} & \multicolumn{6}{c}{\textbf{CWQ}}                                                       \\ \cline{2-7} 
    \multicolumn{1}{c}{} & \textbf{Conj} & \textbf{Compo} & \textbf{Compa} & \textbf{Super} & \textbf{Overall} & \textbf{Ave Price} \\ \hline
    Human Anno & 1.38 / 12.64  & 1.10 / 11.32  & 1.28 / 11.24  & 1.62 / 11.16  & 1.34 / 11.59  & \$0.28 \\ \hline
    Ours w/GPT-4 & 80.67 / 15.25 & 90.67 / 13.16 & 72.00 / 15.11 & 78.67 / 14.75 & 80.50 / 14.57 & \$0.49 \\
    Ours w/GPT-4 + Golden & 78.00 / 13.55 & 91.33 / 11.40 & 75.33 / 13.59 & 85.33 / 12.44 & 82.50 / 12.74 & \$0.38 \\
    Ours w/FT Mistral-7B & 49.33 / 14.71 & 60.00 / 13.69 & 68.67 / 12.71 & 80.67 / 12.05 & 64.67 / 13.29 & -      \\ \hline
    \end{tabular}
}
\caption{Comparison of turn count and cost across on CWQ by question type.}
\label{app_tab:info_dialog_turn_full_cwq}
\end{table*}

\begin{table*}[ht]
\centering
\scalebox{0.8}{
    \begin{tabular}{lcccccc}
    \hline
    \multicolumn{1}{c}{\multirow{2}{*}{\textbf{Method}}} & \multicolumn{6}{c}{\textbf{KQA Pro}} \\ \cline{2-7} 
    \multicolumn{1}{c}{} &
      \textbf{Ct} &
      \textbf{QA} &
      \textbf{QAQ} &
      \textbf{QN} &
      \textbf{QR} &
      \textbf{QRQ} \\ \hline
    Human Anno & 0.86 / 11.28  & 0.74 / 10.76  & 1.16 / 14.04  & 0.96 / 11.12  & 0.30 / 10.44   & 0.80 / 14.04  \\ \hline
    Ours w/GPT-4 & 96.00 / 11.15 & 94.00 / 11.57 & 89.00 / 13.11 & 86.00 / 10.30 & 100.00 / 11.00 & 90.00 / 13.80 \\
    Ours w/FT Mistral-7B & 94.00 / 10.36 & 73.00 / 10.70 & 75.00 / 12.63 & 64.00 / 10.03 & 85.00 / 10.20  & 69.00 / 13.38 \\ \hline
    \multicolumn{1}{c}{(continued)} &
      \textbf{SA} &
      \textbf{SB} &
      \textbf{Vf} &
      \textbf{Overall} &
      \textbf{Ave Price} &
       \\ \hline
    Human Anno & 0.28 / 10.72  & 0.90 / 12.64  & 1.20 / 10.68  & 0.80 / 11.75  & \$0.23         & \\ \hline
    Ours w/GPT-4 & 93.00 / 11.82 & 99.00 / 13.83 & 98.00 / 10.61 & 93.89 / 11.91 & \$0.33         &               \\
    Ours w/FT Mistral-7B & 83.00 / 10.78 & 96.97 / 12.35 & 91.00 / 10.36 & 81.20 / 11.18 & -              &               \\ \hline
    \end{tabular}
}
\caption{Comparison of turn count and cost across on KQA Pro by question type.}
\label{app_tab:info_dialog_turn_full_kqa}
\end{table*}

\begin{table*}[ht]
\centering
    \begin{tabular}{lccccc}
    \hline
    \multicolumn{1}{c}{\multirow{2}{*}{\textbf{Method}}} & \multicolumn{5}{c}{\textbf{MetaQA}}                                 \\ \cline{2-6} 
    \multicolumn{1}{c}{} & \textbf{1-hop} & \textbf{2-hop} & \textbf{3-hop} & \textbf{Overall} & \textbf{Ave Price} \\ \hline
    Human Anno & 0.02 / 6.88  & 0.12 / 7.22   & 0.10 / 7.28  & 0.08 / 7.13  & \$0.06 \\ \hline
    Ours w/GPT-4 & 99.67 / 7.15 & 100.00 / 7.14 & 99.33 / 8.08 & 99.67 / 7.45 & \$0.14 \\
    Ours w/FT Mistral-7B & 99.00 / 7.01 & 99.33 / 7.11  & 98.00 / 7.10 & 98.78 / 7.07 & -      \\ \hline
    \end{tabular}
\caption{Comparison of turn count and cost across on MetaQA by question type.}
\label{app_tab:info_dialog_turn_full_metaqa}
\end{table*}

% add (Reviewer mHdD 2)
\subsection{Impact of Interaction History Length}

\begin{figure*}[htb]
\centering
  \includegraphics[width=0.7\textwidth,page=1]{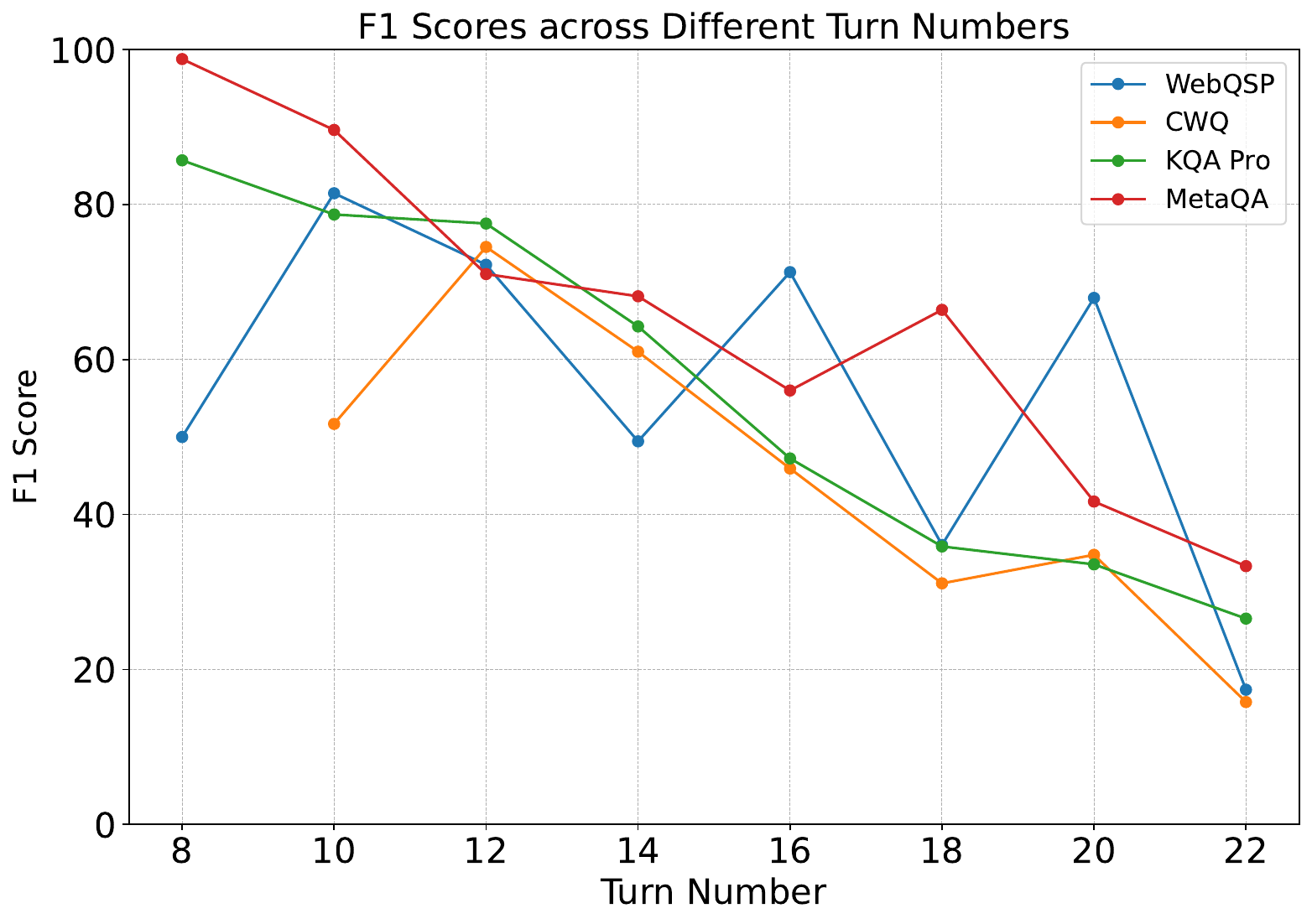}
\caption{F1 Scores across Different Turn Numbers.}
\label{app_fig:f1_score_accross_turns}
\end{figure*}

To investigate the impact of interaction history length on end-to-end performance, we analyzed the changes in F1 score as the number of dialogue turns increased, as presented in Figures \ref{app_fig:f1_score_accross_turns}. It was observed that as the number of dialogue turns increased, the model's performance generally deteriorated.

\subsection{Confusion Matrix of Classifiers}

\begin{figure*}[htb]
\centering
  \includegraphics[width=0.47\textwidth,page=1]{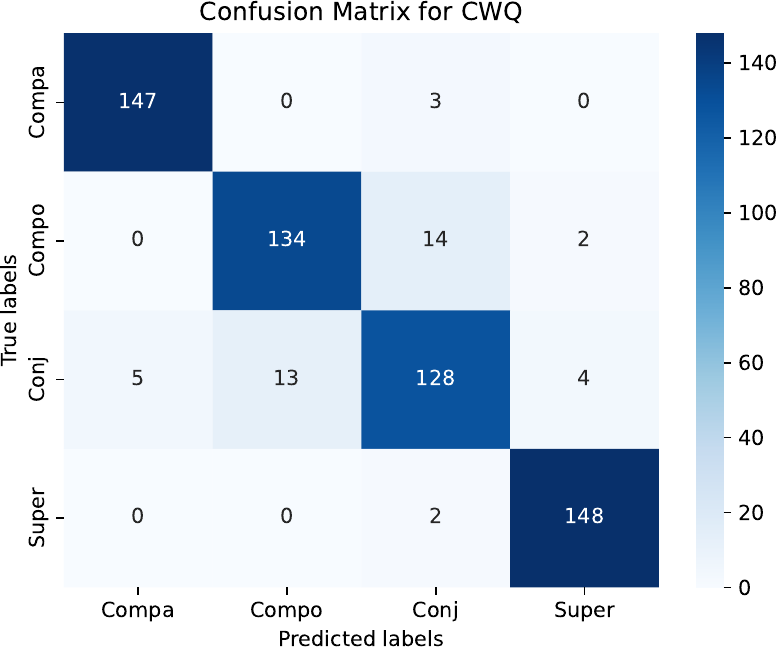}
\hfill
  \includegraphics[width=0.47\textwidth,page=1]{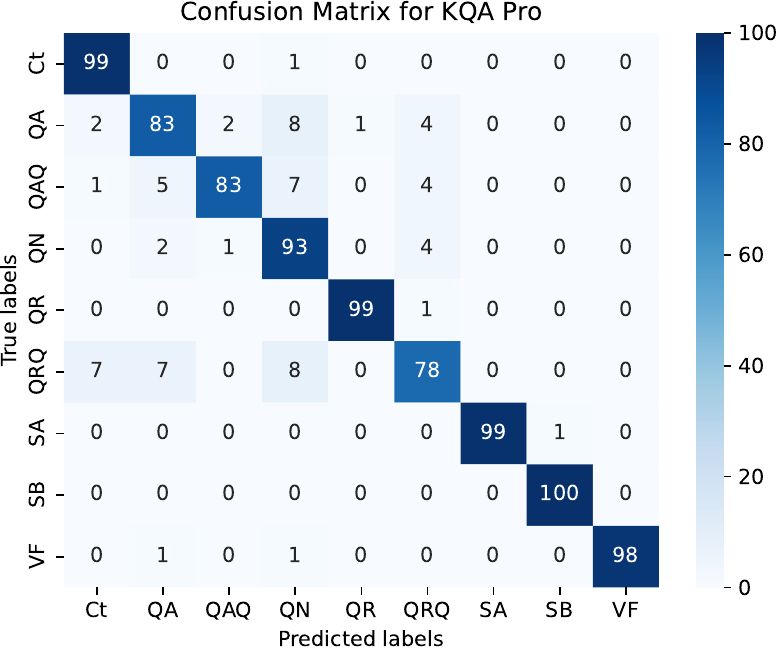}
\caption{Confusion matrix of the classifier for CWQ (left) and KQA Pro (right).}
\label{app_fig:confusion_matrix_cwq_kqa}
\end{figure*}

Table \ref{app_fig:confusion_matrix_cwq_kqa} illustrate the confusion matrices for question type classifiers based on \texttt{bert-base-uncased}.

\subsection{Case Study}
\label{app_sec:case_study}

\begin{figure*}
  \centering
  \includegraphics[width=0.95\textwidth,page=1]{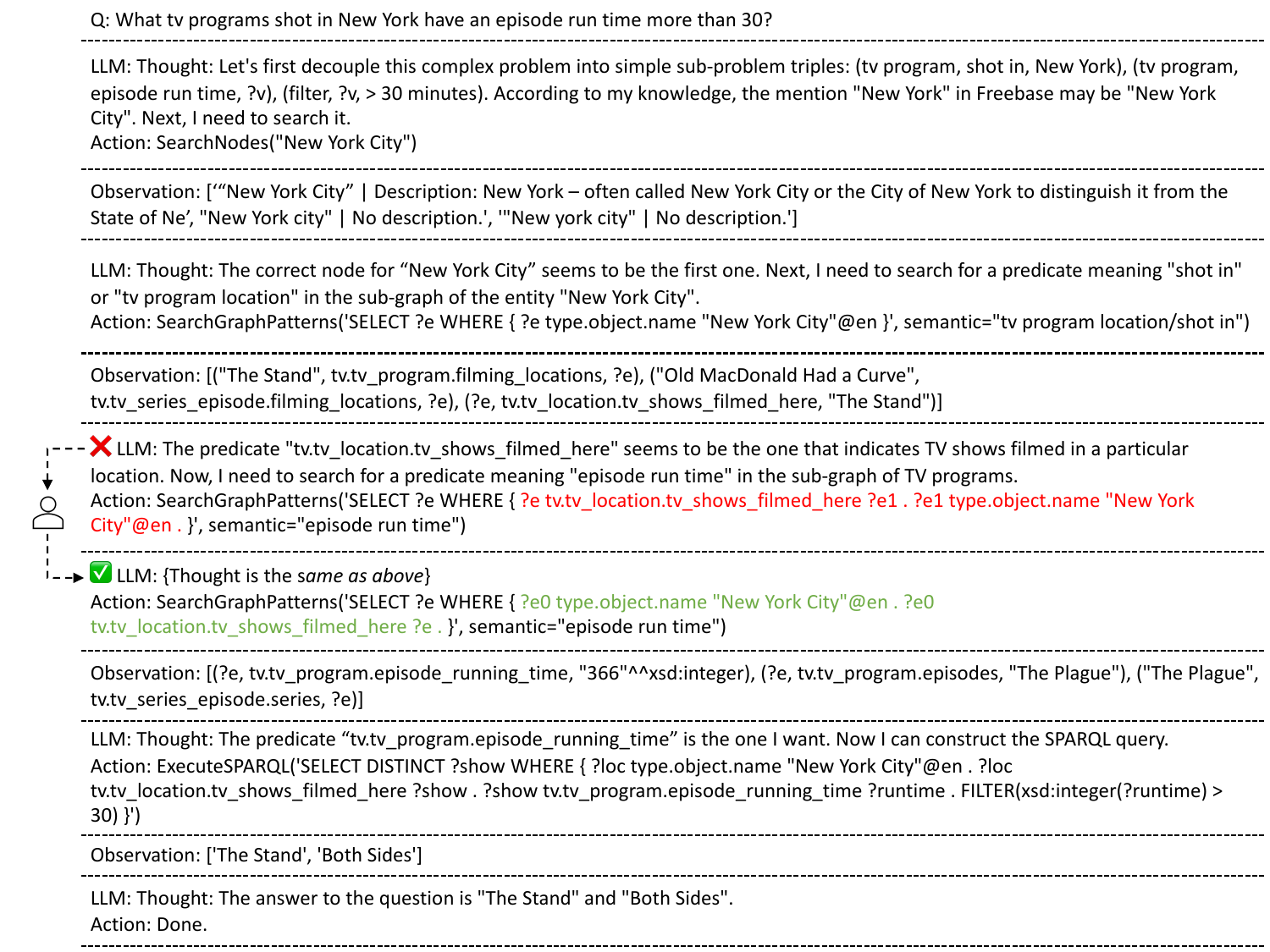}
  \caption{Example of a reasoning error: Human assistance in understanding SPARQL logic.}
  \label{app_fig:case_01_logic_err}
\end{figure*}

\begin{figure*}
  \centering
  \includegraphics[width=0.95\textwidth,page=2]{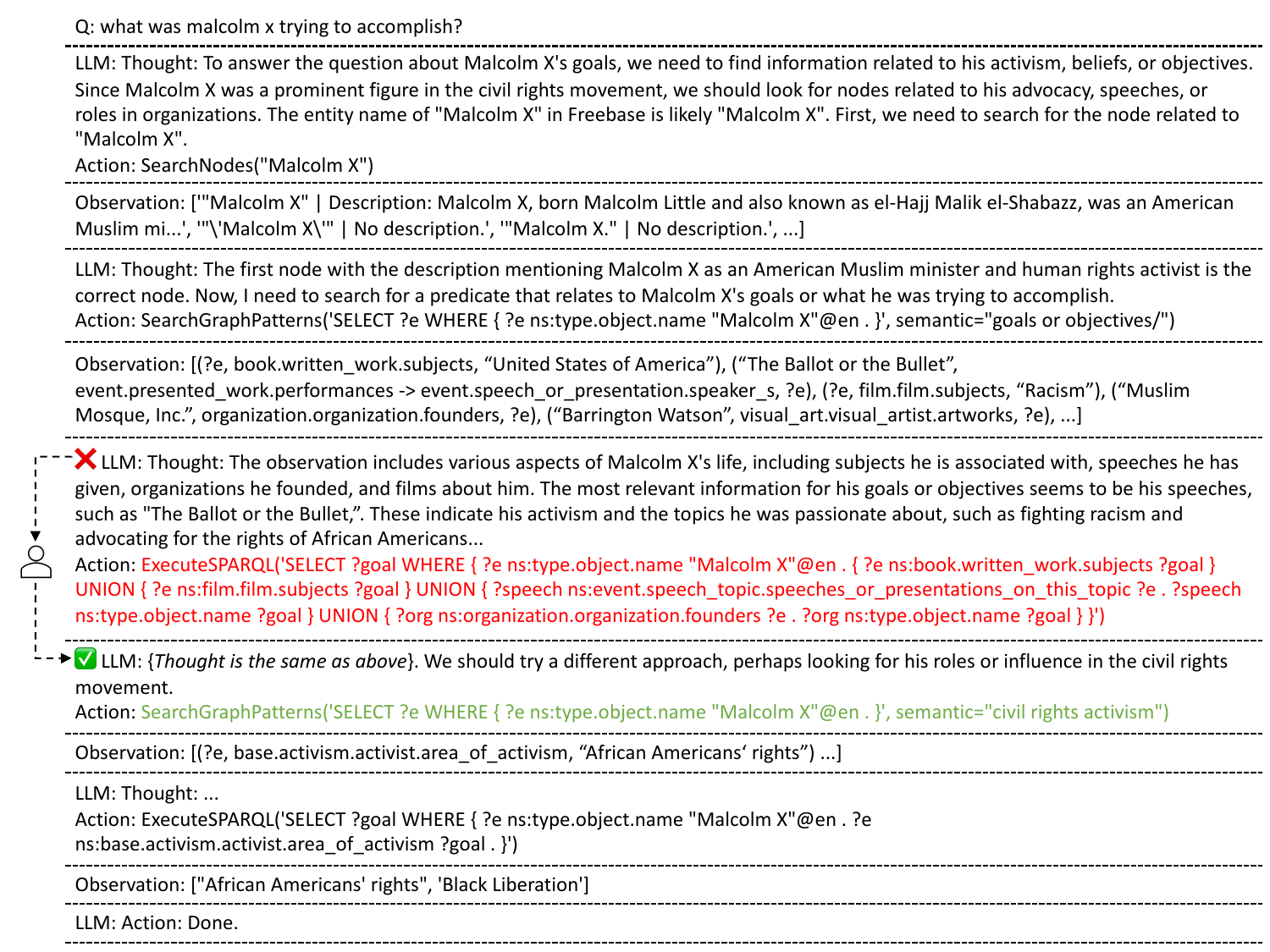}
  \caption{Example of a predicate search error: Human assistance in identifying the predicate.}
  \label{app_fig:case_02_search_p}
\end{figure*}

\begin{figure*}
  \centering
  \includegraphics[width=0.95\textwidth,page=3]{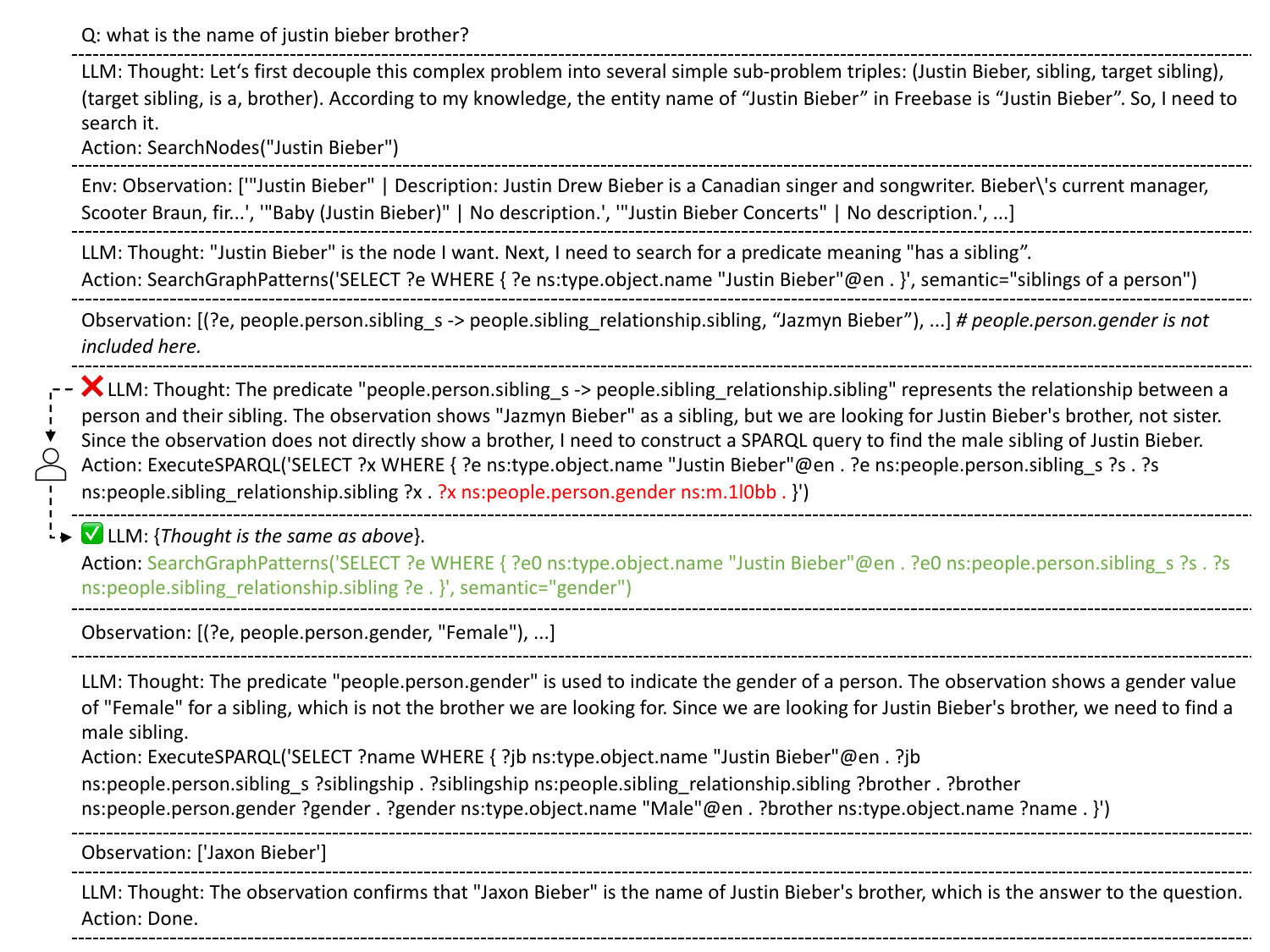}
  \caption{Example of a hallucination error: Human intervention to rectify the hallucination.}
  \label{app_fig:case_03_hallucination}
\end{figure*}

\begin{figure*}
  \centering
  \includegraphics[width=0.95\textwidth,page=4]{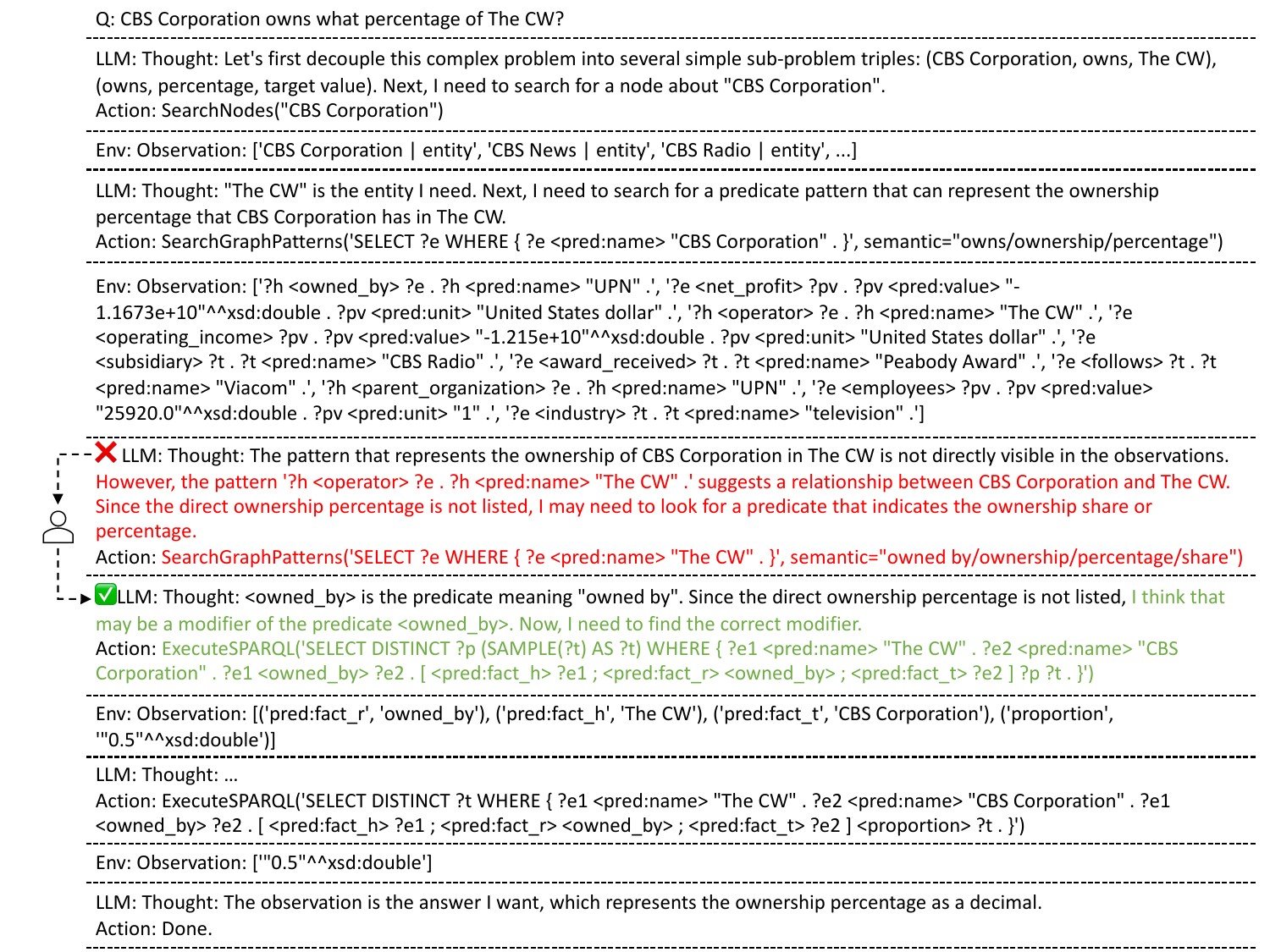}
  \caption{Example of a reasoning error: Human guidance in locating the qualifier.}
  \label{app_fig:case_04_qualifier}
\end{figure*}

In Figure \ref{app_fig:case_01_logic_err}, \ref{app_fig:case_02_search_p}, \ref{app_fig:case_03_hallucination}, and \ref{app_fig:case_04_qualifier}, we present four cases. Each illustrates a error and delineates the correction process with human assistance.

\end{document}